\documentclass{article}

\usepackage{microtype}
\usepackage{graphicx}
\usepackage{subfigure}
\usepackage{booktabs} 
\usepackage{hyperref}

\usepackage[accepted]{icml2025}

\makeatletter
\renewcommand{\ICML@appearing}{%
  \textit{Proceedings of the ICML 2025 Workshop on NewInML, Vancouver, Canada.}%
}
\@ifundefined{ICML@appearingdraft}{}{%
  \renewcommand{\ICML@appearingdraft}{%
    \textit{Proceedings of the ICML 2025 Workshop on NewInML, Vancouver, Canada.}%
  }%
}
\makeatother

\usepackage{amsmath}
\usepackage{amssymb}
\usepackage{mathtools}
\usepackage{amsthm}

\usepackage[utf8]{inputenc}
\usepackage{array}
\usepackage[export]{adjustbox}
\usepackage{arydshln} 
\usepackage{hhline}
\usepackage{multirow}

\usepackage[capitalize,noabbrev]{cleveref}

\usepackage[symbol]{footmisc}
\theoremstyle{plain}

\theoremstyle{definition}

\theoremstyle{remark}

\icmltitlerunning{Coarse-to-Fine Personalized LLM Impressions for Radiology Reports}
 
\begin{document}

\twocolumn[
\icmltitle{Coarse-to-Fine Personalized LLM Impressions for Streamlined Radiology Reports}



\icmlsetsymbol{equal}{*}

\begin{icmlauthorlist}
\icmlauthor{Chengbo Sun}{equal,uchicago}
\icmlauthor{Hui Yi Leong}{equal,uchicago}
\icmlauthor{Lei Li}{uwash} 
\end{icmlauthorlist}

\icmlaffiliation{uchicago}{University of Chicago} 
\icmlaffiliation{uwash}{University of Washington}

\icmlcorrespondingauthor{Chengbo Sun, Hui Yi Leong, Lei Li}
{chengbo1@uchicago.edu, yuki.leong@uchicago.edu, lenny.lilei.cs@gmail.com}

\icmlkeywords{Large Language Models, Radiology, Summarization, Personalization, Reinforcement Learning}

\vskip 0.3in
]

\printAffiliationsAndNotice{\icmlEqualContribution} 

\begin{abstract}
The manual creation of the "Impression" section in radiology reports is a primary driver of radiologist burnout. To address this, we propose a coarse-to-fine framework that uses open-source Large Language Models (LLMs) to automatically generate and personalize impressions from clinical findings. The system first generates a draft and then refines it using machine learning and Reinforcement Learning from Human Feedback (RLHF) to match individual radiologists' styles and ensure factual accuracy. We will fine-tune LLaMA and Mistral models on a large dataset of reports from the University of Chicago Medicine. Our approach aims to significantly reduce administrative workload and enhance reporting workflows while maintaining high standards of clinical precision.
\end{abstract}

\section{Introduction}

This research explores the application of large language models (LLMs) ~\cite{li2025human,cai2024role,yao2025countllm} to a broader range of domains. Specifically, we investigate their use in summarizing radiology reports, addressing critical challenges in clinical decision-making and workflow efficiency. Artificial Intelligence (AI) and Natural Language Processing (NLP) have become integral to healthcare, enabling advancements in patient care, clinical workflows, and medical research. Applications include automating clinical documentation, such as extracting information from clinical notes or summarizing patient histories, as demonstrated in this research. These tools also enhance clinical decision support systems, improving the overall clinical experience.

\begin{figure}[t]
    \centering
    \includegraphics[width=\columnwidth]{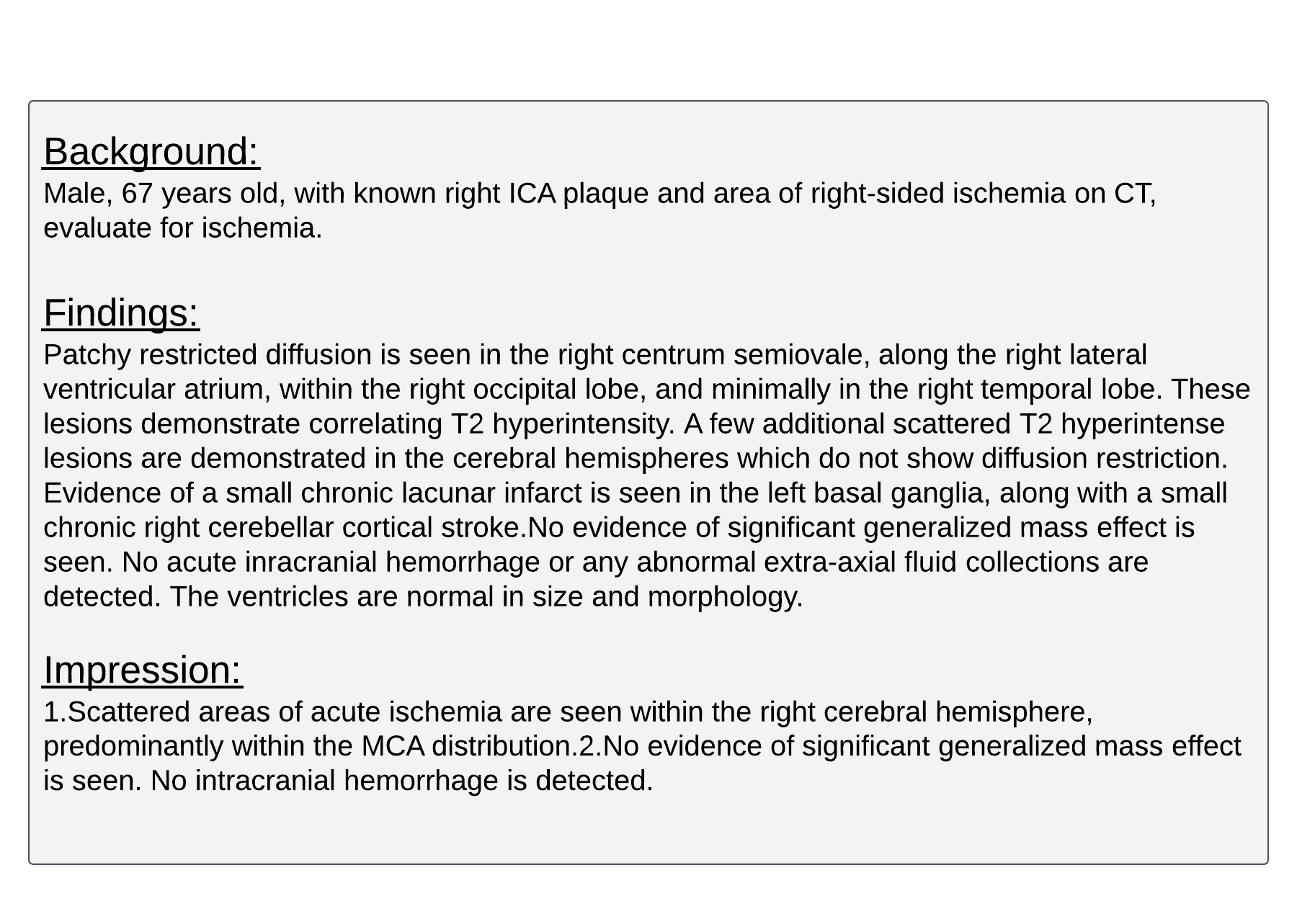}
    \caption{An example of background, finding, and impression.}
    \label{file_example}
\end{figure}

Radiology reports are vital for clinical decision-making; however, generating personalized impressions is a time-intensive task that places a significant burden on radiologists. Radiologists routinely document findings from modalities such as X-rays, CT scans, MRIs, and ultrasounds in free text\cite{gundogdu2021customized}. The Impression section, serving as the core summary, is critical for guiding referring physicians in their decisions. However, crafting impressions is inherently complex and demands a high level of personalization due to the integration of diverse medical data sources and patient-specific requirements. Additionally, the use of domain-specific language, as shown in Figure \ref{file_example}, further adds to the complexity, requiring deep expertise and precision.

Time constraints further exacerbate this challenge, as radiologists often work under tight schedules, leaving limited time to create detailed and personalized reports. Human-written impressions are prone to inefficiencies, errors, and overlooked details. Variability in style, structure, and thoroughness between healthcare providers can lead to discrepancies in report quality, affecting clinical outcomes. Digital tools are increasingly being developed to alleviate this administrative burden, allowing clinicians to focus more on patient care. However, existing LLMs often fail to meet the stylistic and clinical precision required for radiological impressions, highlighting the need for novel techniques to achieve fine-grained personalization.

General-purpose LLMs~\cite{shi2025explaining,cai2025bayesian} are trained on broad datasets that lack the specialized vocabulary, style, and clinical nuances needed for medical reporting. They often cannot adapt their outputs to patient-specific needs or the unique preferences of individual practitioners or institutions\cite{yan2022radbert}. Additionally, these models struggle with maintaining fine-grained control over content and structure, making it difficult to ensure a consistent tone and alignment with input data. Such limitations can lead to trust issues in critical medical contexts.

This research introduces a novel approach to overcoming these challenges by proposing the Coarse-to-Fine generation framework. This framework bridges the gap between generic text generation and the stylistic and clinical precision required for medical imaging reports. It begins with a coarse-grained summary of the findings, capturing essential information from the input data. The outputs are then iteratively refined through fine-grained customization, incorporating patient-specific context, clinical precision, and stylistic alignment with medical standards. Finally, reinforcement learning with human feedback (RLHF) is applied to ensure the outputs are accurate and tailored to the needs of both clinicians and patients.

Our three main contributions are:
\begin{enumerate}
    \item Propose a Coarse-to-Fine generation framework that combines high accuracy summarization with fine-grained personalization and aligns output with clinical and stylistic standards.
    \item Construct and validate a multimodal dataset for personalized patient generation, ensuring the dataset captures diverse patient scenarios and aligns with real-world clinical requirements, enabling robust training and evaluation.
    \item Achieve significant improvements in factual consistency, stylistic alignment, and personalization compared to baseline methods.
\end{enumerate}

\section{Related Work}

\subsection{Domain-Specific Pretrained Models in Biomedical and Clinical NLP}

Early work in clinical NLP successfully adapted language models by pretraining them on domain-specific corpora. This trend began with models like BioBERT \cite{lee2020biobert} on biomedical literature and was extended by BlueBERT \cite{peng2020empirical} and ClinicalBERT \cite{alsentzer2019publicly} using clinical notes from EMRs. These models set new benchmarks on discriminative tasks like named entity recognition (NER) and relation extraction. The approach was further specialized for radiology with RadBERT \cite{yan2022radbert}, which was pretrained on millions of radiology reports for tasks like report coding and summarization. However, a key limitation of these BERT-based, encoder-only architectures is their restricted performance on complex natural language generation (NLG) tasks, which is the primary focus of our work.

\subsection{Foundational models and applications}

The advent of modern foundational LLMs has opened new possibilities for high-quality text generation. However, these models present distinct trade-offs for clinical applications, as summarized in Table \ref{table:LLM_methods}. For example, Mistral-7b \cite{jiang2023mistral} is optimized for inference efficiency, while Gemma models \cite{team2024gemma} are designed for efficient domain adaptation. While proprietary models like GPT-4o \cite{achiam2023gpt} offer exceptional performance, their closed nature limits customization for specific clinical workflows. In contrast, open-source models like Llama-3.1-8B \cite{dubey2024llama} provide a strong balance, supporting long context windows of up to 128K tokens suitable for extensive radiology reports. Given these factors and its superior performance in our preliminary evaluations (Section 5.2), we selected Llama-3.1-8b as the base for our framework.

\subsection{LLM and framework in healthcare}

In the healthcare domain, large language models have first been fine-tuned to generate structured SOAP notes, as demonstrated by Leong et al.’s efficient fine-tuning approach \cite{leong2024efficient,Zhang2025,Zhang2025t}. Building on this work, MediNote is the first end-to-end generative AI framework that combines retrieval-augmented generation (RAG), parameter-efficient fine-tuning, and ambient listening to convert raw clinical conversations directly into SOAP documentation \href{}{\cite{leong2024genai,Wang2025e}}. Complementing these advances,  AgentNet  \cite{leong2024nextgen,} and \cite{wang2025one} employs a fully decentralized, retrieval-augmented architecture in which autonomous agents dynamically reconfigure their roles and interconnections within a directed acyclic graph to optimize clinical workflows and improve patient outcomes in healthcare. Together, these developments chart a clear progression—from LLM fine-tuning for note generation, through RAG-enhanced generative frameworks, to autonomous multi-agent systems—highlighting the evolving landscape of AI-driven healthcare solutions. 

\begin{table}[t]
\caption{Existing LLM methods for Summarization.}
\label{table:LLM_methods}
\vskip 0.15in
\begin{center}
\begin{small}
\begin{sc}
\begin{tabular}{p{2cm} p{2.2cm} p{2.2cm}}
\toprule
\textbf{Method} & \textbf{Advantages} & \textbf{Disadvantages} \\
\midrule
Gemma-2-9b \cite{team2024gemma} &  Offers a trade-off between model size and task efficiency, suitable for diverse NLP tasks. & Performance lag behind more advanced models like Llama 3.1 or GPT-4 in complex reasoning tasks. \\
\hline
Mistral-7b \cite{jiang2023mistral} & Effective for handling large datasets and extended sequences with minimal latency & Require additional fine-tuning for domain-specific tasks like radiology summarization. \\ 
\hline
GPT-4o \cite{achiam2023gpt} & Excels in understanding, summarizing, and reasoning, outperforming most open-source models. & Access requires a subscription, and fine-tuning is often limited or expensive\cite{jiang2024gpt}. \\ 
\hline
Llama-3.1-8B \cite{dubey2024llama} & Supports context windows of up to 128K tokens, suitable for processing extensive radiology reports.Optimized for fewer training samples, reducing computational costs. & Despite its optimizations, fine-tuning and deployment may still demand significant resources. \\ 
\bottomrule
\end{tabular}
\end{sc}
\end{small}
\end{center}
\vskip -0.1in
\end{table}

\section{Data exploration}

The data we use are the 957,134 de-identified radiology reports obtained from the University of Chicago (UC) Medicine\cite{gundogdu2021customized}, curated over the last 12 years until January 1st, 2020. The dataset presented is a collection of radiology reports structured for clinical summarization tasks, making it highly suitable for fine-tuning large language models (LLMs) in the medical domain. It is organized into three primary columns: \textit{clinical information}, \textit{findings}, and \textit{impression}. The clinical information column provides a brief overview of the patient’s medical history and the reason for the radiological examination. The findings column contains detailed imaging observations, while the impression column provides a concise diagnostic summary based on the findings. This structure creates a clear alignment between input (findings) and target output (impression), which is ideal for supervised summarization tasks.

A random sampling of 7,893 reports was analyzed, with an average token count of 123 and a total of approximately 97 million tokens. The Info/Impression Token Count Ratio highlights the difference in length between findings and impressions, with a mean ratio of 7.57, emphasizing the challenge of generating highly concise summaries from more detailed observations.

From a natural language processing (NLP) perspective, the dataset is well-suited for sequence-to-sequence tasks due to its consistent structure and domain specificity. The hierarchical nature of the data—where detailed multi-level observations in the findings are summarized into condensed impressions—poses a unique challenge for LLMs, requiring both factual accuracy and syntactic fluency. The length asymmetry between the findings and impressions adds complexity, as models must process long-context inputs while generating concise outputs without losing critical information.

\section{Methodology}
\subsection{Overview}
\begin{figure*}[t]
\centering
\includegraphics[width=0.9\textwidth]{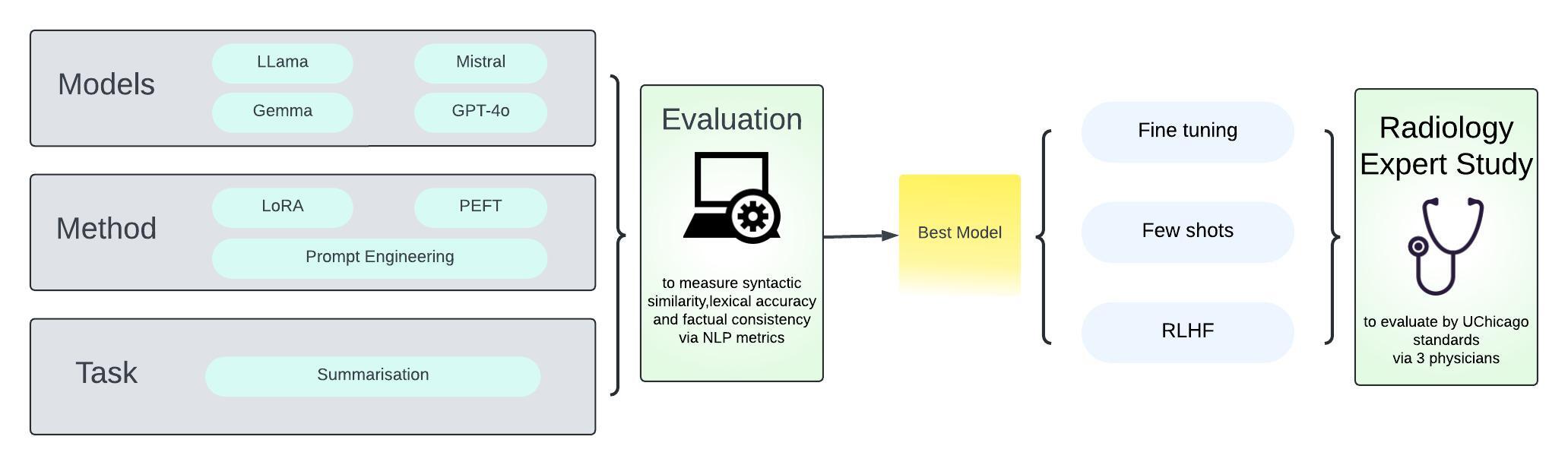}
\caption{\textbf{Architecture Overview:} This architecture outlines the process used for clinical report summarization and evaluation. Multiple large language models (LLMs), including LLaMA, Gemma, Mistral, and GPT-4o, were tested using different methods such as LoRA and PEFT for parameter-efficient fine-tuning, alongside prompt engineering strategies. The models were evaluated on metrics like syntactic similarity, lexical accuracy, and factual consistency to identify the best-performing model. After model selection, further fine-tuning and few-shot learning were applied, followed by reinforcement learning with human feedback (RLHF). Finally, a radiology expert study, conducted by three physicians following UChicago standards, validated the clinical relevance and accuracy of the generated summaries.}
\label{fig:architecture}
\end{figure*}

In this study, we propose a systematic approach to evaluate and fine-tune large language models (LLMs) for the task of radiology findings summarization. The methodology is divided into three main stages: Model Selection, Evaluation, and Radiology Expert Study, as illustrated in the architecture diagram Figure \ref{fig:architecture}.

The proposed method leverages a coarse-to-fine generation approach to automate the creation of radiological impressions, incorporating a dual-stage pipeline for preliminary and refined outputs. The system begins with data preparation for classification. The summarization is performed using the Llama 8B model, which generates the impressions. The generated results are evaluated based on ROUGE scores, BLUE scores, BERT scores and factual consistency.

Once the results are deemed acceptable, they are customized using engineered prompts tailored to three target audiences:

\begin{enumerate}
  \item Brief Summarization: Simplified summaries designed for non-English speakers.
  \item Bullet Point Summarization: Summaries with clear, concise insights for quick review.
  \item Comprehensive Summarization: Detailed summaries aimed at experts.
\end{enumerate}This customized output is designed for diverse clinical interactions, enhancing its applicability across various healthcare scenarios. The proposed methodology ensures accuracy and adaptability, making it a valuable tool for radiology workflows. 

\subsection{Model selection}

\begin{table}[t]
\caption{Summary Performance Metrics Averaged Over 10 Radiology Reports}
\label{table:Summary Performance Metrics Averaged Over 10 Radiology Reports}
\vskip 0.15in
\begin{center}
\begin{small}
\begin{sc}
\begin{tabular}{lcccc}
\toprule
Model & ROUGE & BLUE & BERT & FC \\ \hline
Gemma-2-9b  & 0.368   & 0.032  & 1.78   & 0.58    \\ 
Mistral-7b  & 0.371   & 0.031  & 1.76   & \textbf{0.69}   \\ 
Llama-3.1-8b  & \textbf{0.385}   & \textbf{0.037}   & \textbf{1.78}   & 0.68 \\
\bottomrule
\end{tabular}
\end{sc}
\end{small}
\end{center}
\vskip -0.1in
\end{table}

\textit{Evaluation}
The models were evaluated on a randomly selected subset of 10 radiology findings from the University of Chicago Radiology Report Dataset, ensuring the task's relevance to clinical workflows. The following metrics were used to measure performance:

\begin{itemize}
    \item ROUGE: We applied ROUGE-1, ROUGE-2 and ROUGE-L to quantify syntactic similarity between generated and reference impression, capturing overlapping words and phrases. Formally, ROUGE-N is an n-gram recall between a candidate summary and a set of reference impressions\cite{lin-2004-rouge}. ROUGE-N is computed as follows: 
\[
\text{ROUGE-N} = 
\frac{
\sum_{S \in \{\text{Ref}\}} \sum_{\text{gram}_n \in S} \text{Count}_{\text{match}}(\text{gram}_n)
}{
\sum_{S \in \{\text{Ref}\}} \sum_{\text{gram}_n \in S} \text{Count}(\text{gram}_n)
}
\label{eq:rouge-n}
\tag{1}
\]
Where $n$ stands for the length of the n-gram, gramn, and Countmatch(gramn) is the maximum number of n-grams co-occurring in a generated summary and reference impression.
 We propose using the longest common subsequence LCS-based Fmeasure to estimate the similarity between generated summary $X$ of length $m$ and reference impression $Y$ of length $n$, assuming $X$ is a reference summary sentence and $Y$ is a candidate summary sentence, as follows:
\begin{align}
R_{\text{LCS}} &= \frac{\text{LCS}(X, Y)}{m} \label{eq:lcs-recall} \tag{2} \\
P_{\text{LCS}} &= \frac{\text{LCS}(X, Y)}{n} \label{eq:lcs-precision} \tag{3} \\
F_{\text{LCS}} &= \frac{(1 + b^2) R_{\text{LCS}} P_{\text{LCS}}}{R_{\text{LCS}} + b^2 P_{\text{LCS}}} \label{eq:lcs-f1} \tag{4}
\end{align}

Where $LCS(X,Y)$ is the length of a longest common subsequence of $X$ and $Y$. We call the LCS-based F-measure, i.e. Equation \ref{eq:lcs-f1}, ROUGE-L.

    \item BLEU\cite{Papineni2002BleuAM}: Evaluates lexical accuracy by comparing generated summary n-gram overlap with the reference impression. We first compute the geometric average of the generated summary n-gram precisions, $p_m$ which is the same Equation \ref{eq:rouge-n} using generated summary n-grams up to length $M$ and positive weights $w_m$ summing to one. Next, let $c$ be the length of the candidate translation and $r$ be the effective reference corpus length. We compute the brevity penalty $BP$,

\begin{align}
\text{BP} &= 
\begin{cases} 
1 & \text{if } c > r, \\
e^{(1 - r/c)} & \text{if } c \leq r.
\end{cases} \label{eq:bp} \tag{5}
\end{align} 
Then,
\begin{align}
\text{BLEU} &= \text{BP} \cdot \exp \left( \sum_{m=1}^{M} w_m \log p_m \right) \label{eq:bleu} \tag{6}
\end{align}
    
    \item BERTScore\cite{zhang2020bertscoreevaluatingtextgeneration}: Assesses semantic similarity by comparing embeddings of generated and reference texts, focusing on meaning preservation. We use the tokenizer provided with our model. 
Given a tokenized reference impression sentence $x = \langle x_1, \dots, x_k \rangle$, 
the embedding model generates a sequence of vectors  $\langle \mathbf{x}_1, \dots, \mathbf{x}_k \rangle$. 
Similarly, the tokenized generated summary $\hat{x} = \langle \hat{x}_1, \dots, \hat{x}_m \rangle$ 
is mapped to $\langle \hat{\mathbf{x}}_1, \dots, \hat{\mathbf{x}}_m \rangle$. 
The cosine similarity of a reference impression token $x_i$ and a generated summary token $\hat{x}_j$ is 
$
\frac{\mathbf{x}_i^\top \hat{\mathbf{x}}_j}{\|\mathbf{x}_i\| \|\hat{\mathbf{x}}_j\|}.
$
We use pre-normalized vectors, which reduces this calculation to the inner product $\mathbf{x}_i^\top \hat{\mathbf{x}}_j$. 

The complete score matches each token in $x$ to a token in $\hat{x}$ to compute recall, and each token in $\hat{x}$ to a token in $x$ to compute precision. We use greedy matching to maximize the matching similarity score, where each token is matched to the most similar token in the other sentence. We combine precision and recall to compute an F1 measure. For a reference $x$ and candidate $\hat{x}$, the recall, precision, and F1 scores are:

\begin{equation}
R_{\text{BERT}} = \frac{1}{|x|} \sum_{x_i \in x} \max_{\hat{x}_j \in \hat{x}} x_i^\top \hat{x}_j \tag{7}
\end{equation} 

\begin{equation}
P_{\text{BERT}} = \frac{1}{|\hat{x}|} \sum_{\hat{x}_j \in \hat{x}} \max_{x_i \in x} x_i^\top \hat{x}_j \tag{8}
\end{equation} 

\begin{equation}
F_{\text{BERT}} = \frac{2 \cdot P_{\text{BERT}} \cdot R_{\text{BERT}}}{P_{\text{BERT}} + R_{\text{BERT}}} \tag{9}
\end{equation}
    
    \item Factual Consistency (FC)\cite{kryscinski2019evaluating}: Measures alignment between generated summaries and input findings to ensure the generated output accurately reflects the clinical observations. To evaluate the factual consistency of our model-generated summaries, we applied the TrueTeacher-based T5-11B model\cite{gekhman2023trueteacher}, which was fine-tuned on ANLI and TrueTeacher data. This model was used as a factual evaluator to compare generated summaries against the original findings from radiology reports. We framed the task as a Natural Language Inference (NLI) problem, where the generated summary served as the hypothesis and the original findings as the premise. The model provided entailment probabilities that indicated the degree of factual consistency, offering a quantitative measure of how accurately the generated summary preserved the key information from the original findings. This methodology allowed for robust, scalable evaluation of factual consistency in a domain-specific context. 
\end{itemize}

Each metric highlights different aspects of summarization: syntax, semantics, lexical accuracy, and factual consistency. Together, they provide a holistic view of model performance.

As shown in Table \ref{table:Summary Performance Metrics Averaged Over 10 Radiology Reports}, Llama-3.1-8b consistently outperformed the other models across all metrics. It achieved the highest ROUGE score of 0.385, indicating superior overlap between its generated summaries and the ground truth impressions. Its BLEU score of 0.037 demonstrates better word-level precision compared to Gemma-2-9b and Mistral-7b. The BERTScore of 1.78 further highlights Llama-3.1-8b's contextual and semantic alignment with the reference summaries, showcasing its ability to capture nuanced medical terminology.

In addition to linguistic quality, factual consistency is critical for radiology findings summarization. Mistral-7b achieved a factual consistency score of 0.69, narrowly outperforming Llama-3.1-8b at 0.68. Despite this minor difference, Llama-3.1-8b provided the most balanced performance across all metrics, demonstrating its robustness in handling radiology-specific language and maintaining factual integrity.

Our results confirm that Llama-3.1-8b is the most suitable choice for radiology findings summarization due to its superior linguistic quality, semantic understanding, and balanced factual consistency. These results underscore the model’s capability to address the unique challenges of medical summarization tasks, such as handling domain-specific terminology and ensuring clinical relevance. 

\textit{Best Model Selection and Fine-Tuning}
Based on the evaluation, LLaMA-3.1-8b emerged as the best model, achieving the highest ROUGE, BLEU, and BERTScore, with a factual consistency score of 0.68. This model was further optimized through:

Fine-Tuning: The model was trained on additional domain-specific radiology datasets to align with clinical language and terminology.
Few-Shot Learning: Leveraging few-shot prompts, the model was fine-tuned to generalize effectively with minimal examples.
Reinforcement Learning with Human Feedback (RLHF): Incorporating feedback from radiologists to refine the generated summaries, ensuring clinical relevance and coherence.
These techniques improved the model’s ability to generate summaries that are both accurate and clinically useful.

\subsection{Model Training and Fine-tuning}
The proposed approach utilizes a parameter-efficient fine-tuning (PEFT) technique implemented with the FastLanguageModel library\cite{xu2023parameter}. The base model is optimized using Low-Rank Adaptation (LoRA)\cite{hao2024flora}, a strategy that adds trainable low-rank matrices to specific layers of the model, enabling efficient adaptation to new tasks with minimal computational overhead. We fine-tuned the base model using parameter-efficient fine-tuning (PEFT) with LoRA, implemented via the FastLanguageModel library. A rank of 16 was chosen to balance efficiency and performance, with adapters applied to core transformer modules. LoRA alpha was set to 16, with dropout disabled to fully utilize gradient updates. Bias was set to "none" to reduce parameter overhead. Gradient checkpointing with UnsLoft reduced VRAM usage by 30\%, enabling longer sequences and larger batch sizes. A fixed random seed (3407) ensured reproducibility. While options like LoftQ and RSLora were not used, they remain viable for future optimization. The model is fine-tuned using Supervised Fine-Tuning (SFT) with the \texttt{SFTTrainer} class, a robust and flexible framework for adapting large language models to custom datasets. The fine-tuning process begins with data preparation, where the training dataset is preprocessed to ensure proper tokenization and sequence length adjustment. Long sequences are truncated to fit the model’s maximum sequence length, and the \texttt{text} field of the dataset is used for training inputs. During training, the batch size is set to 1 per device, with gradient accumulation steps configured as 2 to effectively simulate a batch size of 2. Training is conducted over 50 epochs with a maximum of 60 gradient updates, balancing computational cost and fine-tuning depth. A linear learning rate scheduler is applied, starting at 1e-4, with 5 warmup steps to stabilize the optimization process. Mixed-precision training, leveraging FP16 or BF16 precision based on hardware support, is employed to improve computational efficiency. The AdamW optimizer with 8-bit precision (\texttt{adamw\_8bit}) reduces memory requirements while maintaining performance, and a weight decay of 0.01 is used to prevent overfitting. Logging occurs at every step to monitor progress, and all outputs are stored in the \texttt{outputs} directory for subsequent evaluation. To streamline the training process, external reporting tools such as WandB are disabled. This fine-tuning strategy integrates the parameter efficiency of LoRA and the scalability of the \texttt{SFTTrainer}, enabling the adaptation of pre-trained large language models to domain-specific tasks while minimizing computational and memory overhead. The approach ensures optimal model performance on limited hardware resources, making it suitable for real-world applications.

\subsection{Coarse-to-Fine Generation Process}
\begin{figure*}[t]
    \centering
    \includegraphics[width=0.8\textwidth]{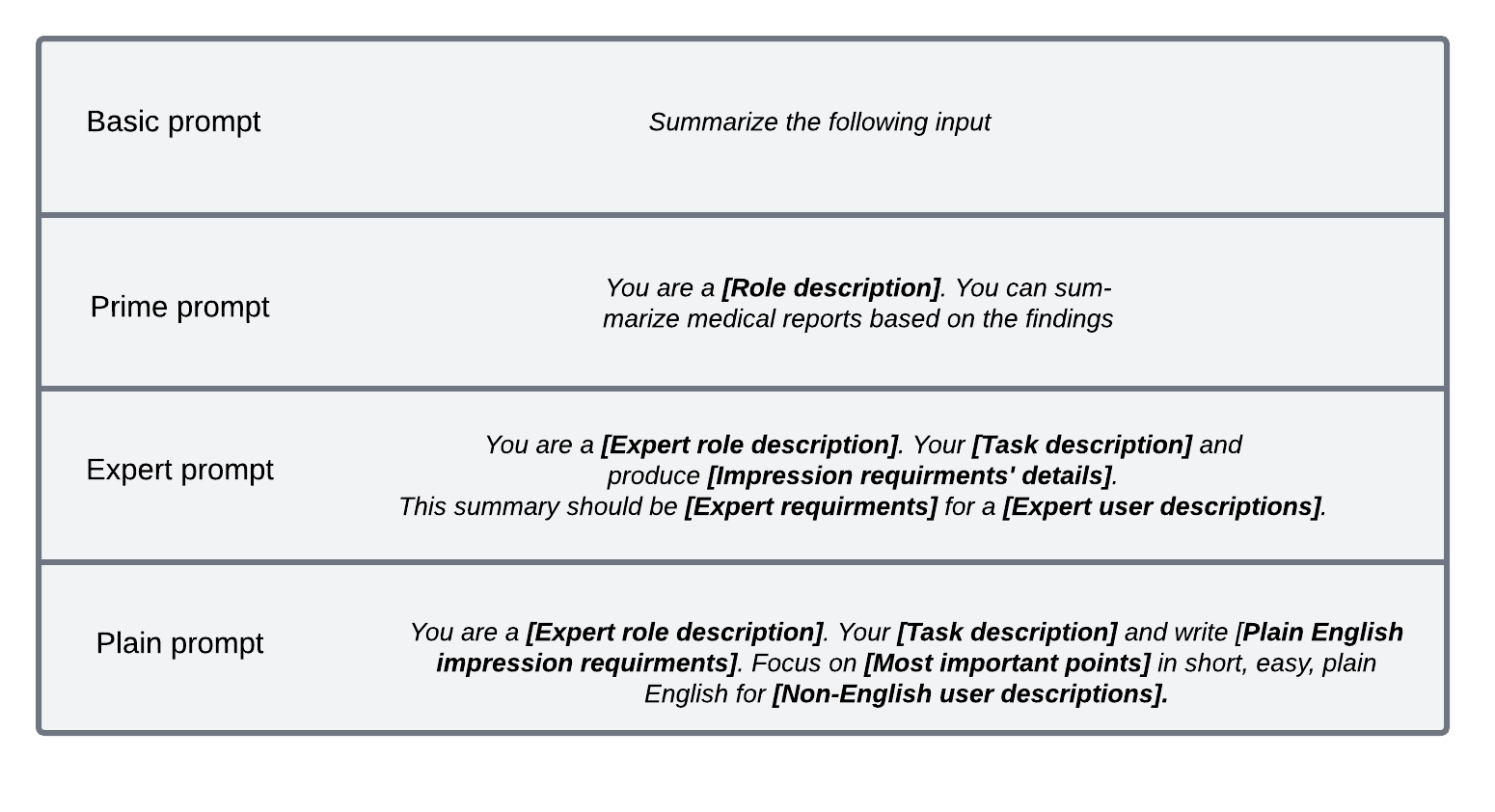}
    \caption{Prompts examples}
    \label{fig:Prompts}
\end{figure*}

The Coarse Stage initiates the summarization process by utilizing large language models (LLMs) to generate preliminary impressions based on the input findings and patient data. This stage is designed to capture primary clinical insights while ensuring the output follows a structured format.
We designed a prompt engineering strategy that evolves from a simple base prompt to more detailed and customized versions tailored for specific audiences. The base model utilizes a straightforward prompt, such as \texttt{"Summarize the following input"}\cite{wang2023plan}, which generates bullet-point key points. Building on this, we incorporated role descriptions and task-specific instructions to produce more detailed outputs while maintaining the bullet-point structure. Additionally, we provided three-shot examples to guide the model in generating outputs aligned with specific writing styles and standards. For expert-level outputs, the prompts further emphasize role descriptions, expanded task instructions, and a clear definition of the target audience to ensure the generated content meets their expectations. Users can also adjust the output length based on specific requirements. Full details of the prompts, including examples for base, detailed, and expert versions, are provided in Figure \ref{fig:Prompts}.

This structured progression not only improves the quality of the generated outputs but also enhances the model's ability to adapt to varying user requirements, ensuring consistency and precision. The flow chart visually encapsulates this iterative and hierarchical prompt engineering strategy, emphasizing the systematic refinement process.

\section{Experiment}
\subsection{Experiment setup}

For our experiments, we curated a training set of 500 reports from the University of Chicago dataset, balanced and stratified by patient gender and age. An additional 50 reports were held out for testing. Our model was fine-tuned using a parameter-efficient approach (PEFT) with Low-Rank Adaptation (LoRA), setting the rank to 16 and alpha to 16. We utilized the SFTTrainer with a batch size of 2 (1 per device with 2 gradient accumulation steps), the AdamW 8-bit optimizer, and a linear learning rate scheduler starting at 1e-4. The model was trained for up to 50 epochs on a single Google T4 GPU, a process which took less than 25 minutes. All evaluation was performed using the metrics detailed in Section 4.2. An exhaustive list of all hyperparameters can be found in the appendix.

\subsection{Comparison experiments.}
\begin{table}[h]
\caption{Summary Performance Metrics Averaged Over 10 Radiology Reports}
\label{table:comparison_results}
\vskip 0.15in
\begin{center}
\begin{small}
\begin{sc}
\begin{tabular}{lcccc}
\toprule
Model & ROUGE & BLUE & BERT & FC \\ \hline
Gemma-2-9b    & 0.368   & 0.032  & 1.78   & 0.58    \\ 
Mistral-7b    & 0.371   & 0.031  & 1.76   & \textbf{0.69}    \\ 
Llama-3.1-8b  & 0.385   & 0.037  & 1.78   & 0.68    \\ \hline
Corse2Fine    & \textbf{0.420}   & \textbf{0.041}   & \textbf{2.51}   & 0.612 \\
\bottomrule
\end{tabular}
\end{sc}
\end{small}
\end{center}
\vskip -0.1in
\end{table}

\begin{figure*}[t]
    \centering
    \includegraphics[width=\textwidth]{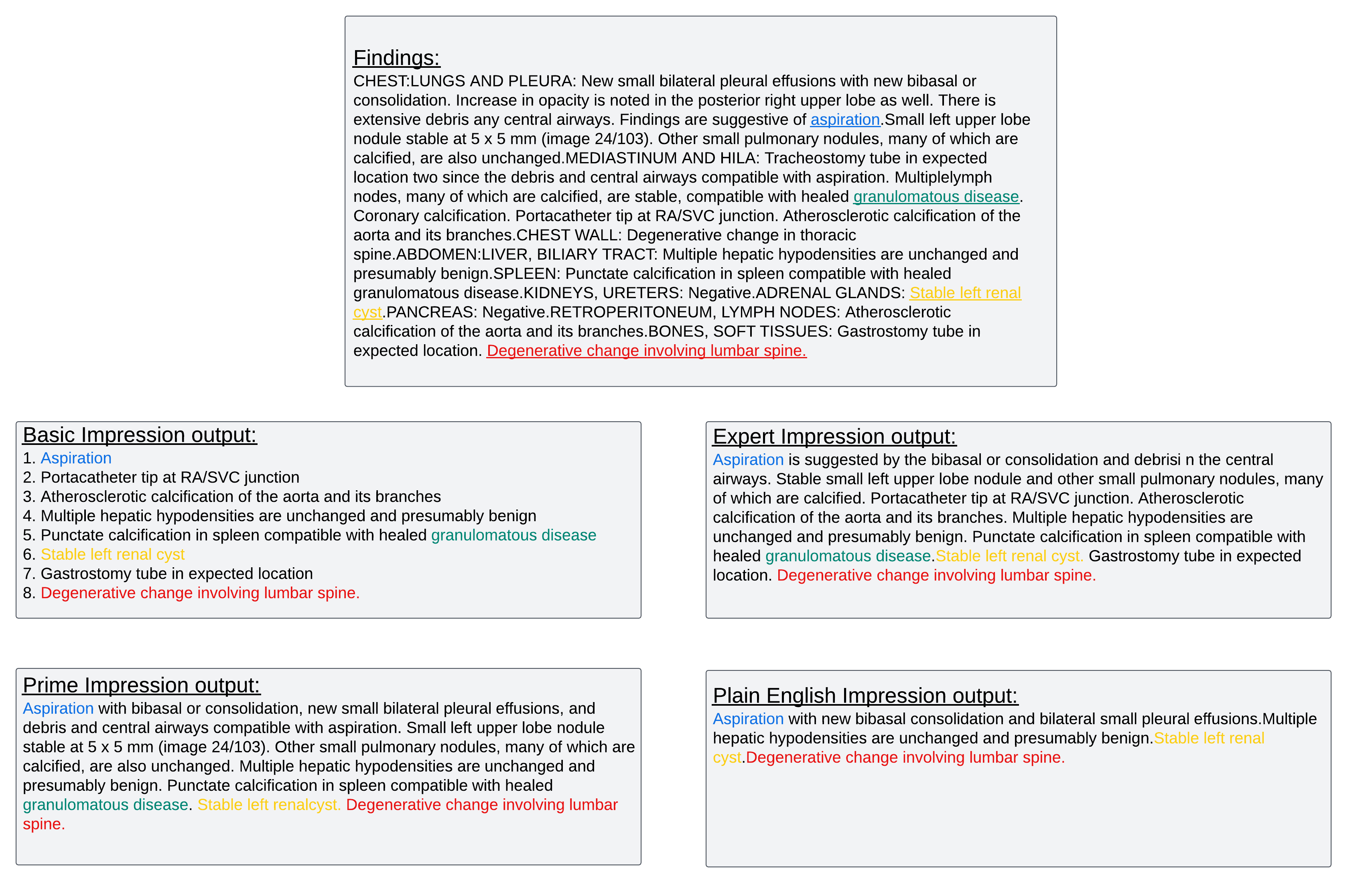}
    \caption{Findings and generated impression examples}
    \label{fig:results examples}
\end{figure*}

\textit{Output results}
We introduced a style-conditioning mechanism that allows the model to generate impressions tailored to user preferences, ranging from concise summaries to more elaborate outputs. Examples of the results are shown in Figure \ref{fig:results examples}. The original finding was notably long, but aside from highlighting the key disease, as shown in Figure \ref{fig:results examples}, it also included information on 10 additional examinations, all of which were either negative or unchanged. Consequently, the generated impression summarized these findings as "Stable examination."

Our model demonstrated the ability to accurately summarize the relevant disease while appropriately disregarding the negative or unchanged results. This highlights the model's effectiveness in prioritizing clinically significant information and producing concise, context-appropriate impressions, improving both efficiency and clarity for end users.

The highlighted sections in the findings represent the most important details. These key points were successfully captured and summarized in all versions of the generated outputs, with accurate and contextually appropriate descriptions.

To address this limitation, we conducted human assessments involving one radiologist from UC Medicine and an independent external board-certified radiologist.

\begin{figure}[h]
    \centering
    \includegraphics[width=\columnwidth]{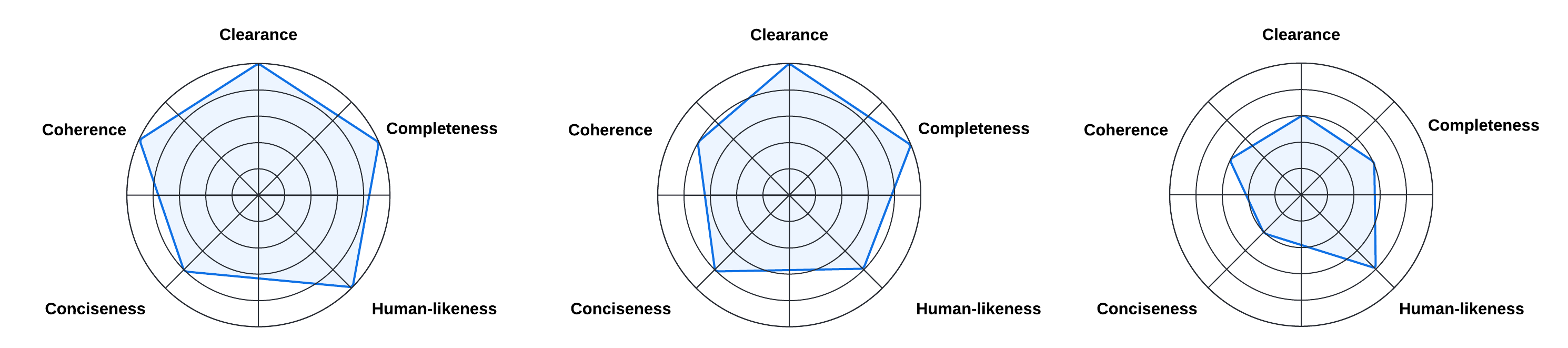}
    \caption{\textbf{Hexagon Rating:} From left to right are the positive rating, neutral rating, and negative rating respectively, illustrating the average rating scores for the 200 generated reports. The pentagon visualizes five key dimensions of summary quality: Clearance, Completeness, Human-likeness, Conciseness, and Coherence. Higher values along each axis indicate better performance in the corresponding metric.}
    \label{fig:rating}
\end{figure}

We randomly selected 300 reports from the UC Medicine test set, consisting of 200 generated by our fine-tuned model and 100 original reports, randomly mixed. Radiologists compared the predicted impressions with the original ones, assigning one of three ratings: 'positive' if the model-generated impression was preferred, 'negative' if the original impression was deemed superior, and 'neutral' if both were equally accurate. The figure\ref{fig:rating} illustrates the average rating scores across five dimensions: Clearance, Completeness, Human-likeness, Conciseness, and Coherence. Notably, 'positive' ratings were frequently given when the model successfully captured incidental findings that were omitted in the original impressions.

Out of the 200 generated reports, 145 were labeled as “neutral,” 14 as “positive,” and 30 as “negative.” Eleven reports were excluded due to unresolved discrepancies between the evaluators, leaving 289 reports for the final analysis. Overall, 79.5\% of the evaluated samples received either “neutral” or “positive” votes, indicating that the majority of the system’s predictions were considered by radiologists to be at least as accurate as human-generated impressions.

\textit{Stability test} To evaluate robustness against real-world data entry errors, we tested our model on findings with a simulated 3\% typographical error rate. The model's performance remained remarkably stable, showing only minimal degradation compared to the clean dataset (ROUGE: 0.370, BERT: 2.49). This resilience to noisy input highlights its reliability for practical clinical applications.

\begin{figure}[t]
    \centering
    \includegraphics[width=\columnwidth]{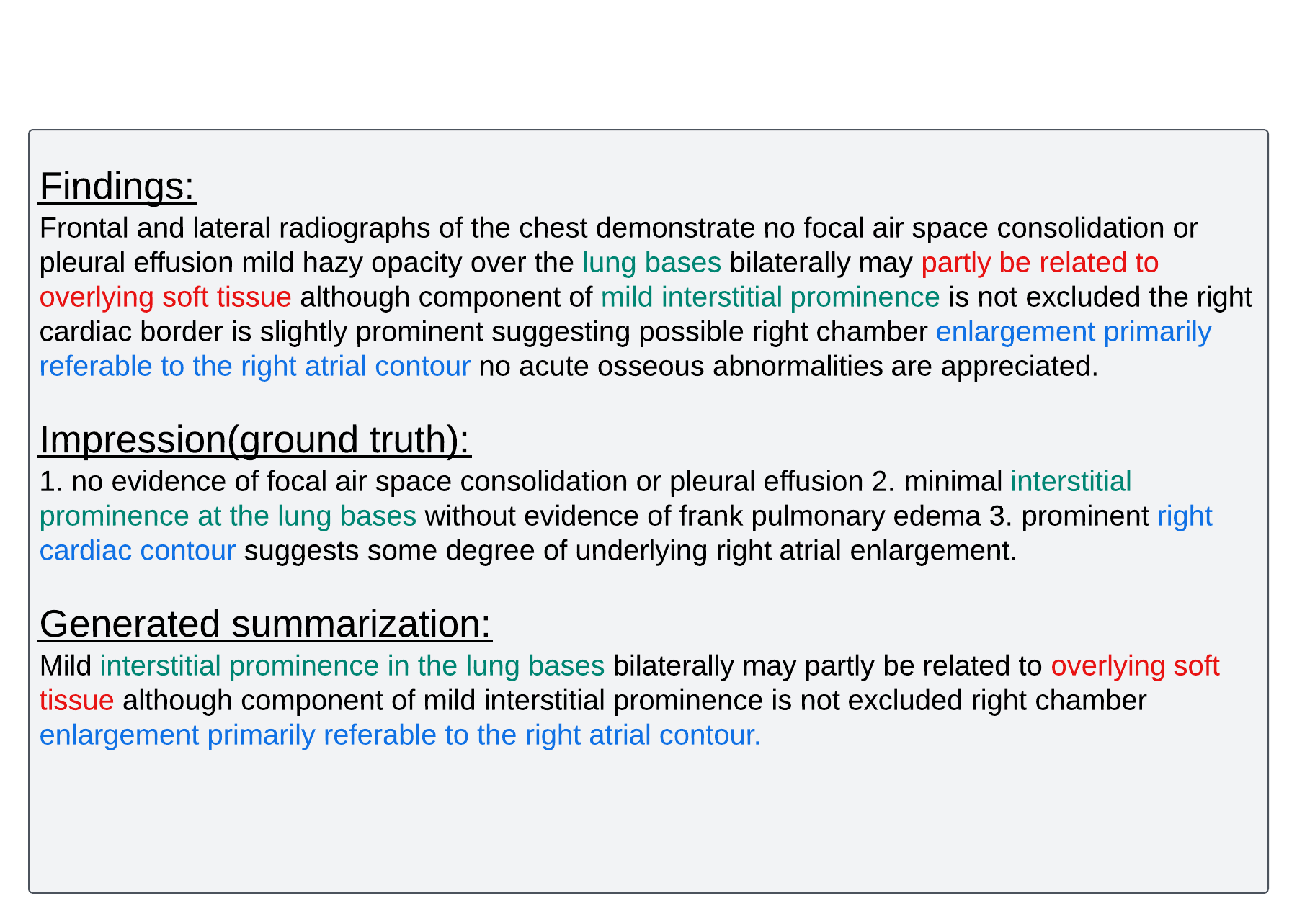}
    \caption{Generated result example of CheXpert Plus dataset}
    \label{fig:Generated results example of CheXpert Plus dataset}
\end{figure}

\textit{Universality test} To test the generalizability of our model, we applied it to the external CheXpert Plus dataset. As shown in the example in Figure \ref{fig:Generated results example of CheXpert Plus dataset}, the model effectively summarized key clinical findings even with a different data source and writing style. In a review of 10 reports from this dataset, radiologists rated the generated impressions as equal to or better than the original in 80\% of cases (3 positive, 5 neutral, 2 negative), demonstrating the model's strong versatility.

Besides, 10 results haven been reviewed by our radiologist, 3 reports are marked positive, 5 reports are marked natural and 2 are marked negative. the results show our model's strong versatility for other institution's dataset or writing styles. 

\section{Conclusion}

In this paper, we presented a Coarse-to-Fine Personalized LLM framework for generating streamlined radiology impressions. Our approach leverages open-source models and parameter-efficient fine-tuning techniques, such as LoRA, to generate clinically relevant, stylistically aligned summaries. Through a two-stage generation process—beginning with coarse summaries and refined via prompt engineering and reinforcement learning with human feedback (RLHF)—the system adapts to diverse clinical needs. Experiments on the University of Chicago dataset demonstrated improvements in factual consistency, linguistic quality, and expert preference ratings. Stability and generalizability were further validated through robustness testing and application to the CheXpert Plus dataset. Looking ahead, we aim to integrate visual data and explore advanced multi-modal models to enhance clinical reasoning and support real-world diagnostic workflows.

\nocite{*}
\bibliographystyle{icml2025}
\bibliography{mybibliography} 

\begin{thebibliography}{121}
\providecommand{\natexlab}[1]{#1}
\providecommand{\url}[1]{\texttt{#1}}
\expandafter\ifx\csname urlstyle\endcsname\relax
  \providecommand{\doi}[1]{doi: #1}\else
  \providecommand{\doi}{doi: \begingroup \urlstyle{rm}\Url}\fi

\bibitem[Achiam et~al.(2023)Achiam, Adler, Agarwal, Ahmad, Akkaya, Aleman, Almeida, Altenschmidt, Altman, Anadkat, et~al.]{achiam2023gpt}
Achiam, J., Adler, S., Agarwal, S., Ahmad, L., Akkaya, I., Aleman, F.~L., Almeida, D., Altenschmidt, J., Altman, S., Anadkat, S., et~al.
\newblock Gpt-4 technical report.
\newblock \emph{arXiv preprint arXiv:2303.08774}, 2023.

\bibitem[Alsentzer et~al.(2019)Alsentzer, Murphy, Boag, Weng, Jin, Naumann, and McDermott]{alsentzer2019publicly}
Alsentzer, E., Murphy, J.~R., Boag, W., Weng, W.-H., Jin, D., Naumann, T., and McDermott, M.
\newblock Publicly available clinical bert embeddings.
\newblock \emph{arXiv preprint arXiv:1904.03323}, 2019.

\bibitem[Cai et~al.(2024)Cai, Zhao, Liu, Jiang, Zhang, Wu, Hwang, Belongie, and Li]{cai2024role}
Cai, C., Zhao, X., Liu, H., Jiang, Z., Zhang, T., Wu, Z., Hwang, J.-N., Belongie, S., and Li, L.
\newblock The role of deductive and inductive reasoning in large language models.
\newblock \emph{arXiv preprint arXiv:2410.02892}, 2024.

\bibitem[Cai et~al.(2025)Cai, Liu, Zhao, Jiang, Zhang, Wu, Lee, Hwang, and Li]{cai2025bayesian}
Cai, C., Liu, H., Zhao, X., Jiang, Z., Zhang, T., Wu, Z., Lee, J., Hwang, J.-N., and Li, L.
\newblock Bayesian optimization for controlled image editing via llms.
\newblock \emph{arXiv preprint arXiv:2502.18116}, 2025.

\bibitem[Chen et~al.(2021)Chen, Lu, Wu, Clarke, Yu, Van~Eyk, Herrington, and Wang]{chen2021data}
Chen, L., Lu, Y., Wu, C.-T., Clarke, R., Yu, G., Van~Eyk, J.~E., Herrington, D.~M., and Wang, Y.
\newblock Data-driven detection of subtype-specific differentially expressed genes.
\newblock \emph{Scientific reports}, 2021.

\bibitem[Chen et~al.(2023)Chen, Xiang, Lu, Xuan, Wang, Chen, and Yang]{chen2023rgp}
Chen, Z., Xiang, J., Lu, Y., Xuan, Q., Wang, Z., Chen, G., and Yang, X.
\newblock Rgp: Neural network pruning through regular graph with edges swapping.
\newblock \emph{IEEE Transactions on Neural Networks and Learning Systems}, 35\penalty0 (10):\penalty0 14671--14683, 2023.

\bibitem[Chu et~al.(2025{\natexlab{a}})Chu, Lin, Xiang, Shen, Su, Chu, Yang, Zhang, Wu, and Zhang]{chu2025selective}
Chu, K., Lin, Z., Xiang, D., Shen, Z., Su, J., Chu, C., Yang, Y., Zhang, W., Wu, W., and Zhang, W.
\newblock Selective kv-cache sharing to mitigate timing side-channels in llm inference.
\newblock \emph{arXiv preprint arXiv:2508.08438}, 2025{\natexlab{a}}.

\bibitem[Chu et~al.(2025{\natexlab{b}})Chu, Shen, Xiang, and Zhang]{chu2025safekv}
Chu, K., Shen, Z., Xiang, D., and Zhang, W.
\newblock Safekv: Safe kv-cache sharing in llm serving.
\newblock In \emph{Machine Learning for Computer Architecture and Systems}, 2025{\natexlab{b}}.

\bibitem[Ding et~al.(2025)Ding, Xiang, Sun, Qi, and Zhao]{ding2025ai}
Ding, T., Xiang, D., Sun, T., Qi, Y., and Zhao, Z.
\newblock Ai-driven prognostics for state of health prediction in li-ion batteries: A comprehensive analysis with validation.
\newblock \emph{arXiv preprint arXiv:2504.05728}, 2025.

\bibitem[Du et~al.(2024)Du, Bhardwaj, Lu, Wang, Parker, Zhang, Van~Eyk, Yu, Clarke, Herrington, et~al.]{du2024embracing}
Du, D., Bhardwaj, S., Lu, Y., Wang, Y., Parker, S.~J., Zhang, Z., Van~Eyk, J.~E., Yu, G., Clarke, R., Herrington, D.~M., et~al.
\newblock Embracing the informative missingness and silent gene in analyzing biologically diverse samples.
\newblock \emph{Scientific reports}, 14\penalty0 (1):\penalty0 28265, 2024.

\bibitem[Dubey et~al.(2024)Dubey, Jauhri, Pandey, Kadian, Al-Dahle, Letman, Mathur, Schelten, Yang, Fan, et~al.]{dubey2024llama}
Dubey, A., Jauhri, A., Pandey, A., Kadian, A., Al-Dahle, A., Letman, A., Mathur, A., Schelten, A., Yang, A., Fan, A., et~al.
\newblock The llama 3 herd of models.
\newblock \emph{arXiv preprint arXiv:2407.21783}, 2024.

\bibitem[Durmus et~al.(2020)Durmus, He, and Diab]{durmus2020feqa}
Durmus, E., He, H., and Diab, M.
\newblock Feqa: A question answering evaluation framework for faithfulness assessment in abstractive summarization.
\newblock \emph{arXiv preprint arXiv:2005.03754}, 2020.

\bibitem[Fu et~al.(2024)Fu, Lu, Wang, Zhang, Zhang, Yu, Liu, Clarke, Herrington, and Wang]{fu2024ddn3}
Fu, Y., Lu, Y., Wang, Y., Zhang, B., Zhang, Z., Yu, G., Liu, C., Clarke, R., Herrington, D.~M., and Wang, Y.
\newblock Ddn3. 0: Determining significant rewiring of biological network structure with differential dependency networks.
\newblock \emph{Bioinformatics}, pp.\  btae376, 2024.

\bibitem[Gekhman et~al.(2023)Gekhman, Herzig, Aharoni, Elkind, and Szpektor]{gekhman2023trueteacher}
Gekhman, Z., Herzig, J., Aharoni, R., Elkind, C., and Szpektor, I.
\newblock Trueteacher: Learning factual consistency evaluation with large language models, 2023.

\bibitem[Gundogdu et~al.(2021)Gundogdu, Pamuksuz, Chung, Telleria, Liu, Khan, and Chang]{gundogdu2021customized}
Gundogdu, B., Pamuksuz, U., Chung, J.~H., Telleria, J.~M., Liu, P., Khan, F., and Chang, P.~J.
\newblock Customized impression prediction from radiology reports using bert and lstm s.
\newblock \emph{IEEE Transactions on Artificial Intelligence}, 4\penalty0 (4):\penalty0 744--753, 2021.

\bibitem[Guo et~al.(2024)Guo, Chen, Wang, Chang, Pei, Chawla, Wiest, and Zhang]{guo2024large}
Guo, T., Chen, X., Wang, Y., Chang, R., Pei, S., Chawla, N.~V., Wiest, O., and Zhang, X.
\newblock Large language model based multi-agents: A survey of progress and challenges.
\newblock \emph{arXiv preprint arXiv:2402.01680}, 2024.

\bibitem[Hao et~al.(2024)Hao, Cao, and Mou]{hao2024flora}
Hao, Y., Cao, Y., and Mou, L.
\newblock Flora: Low-rank adapters are secretly gradient compressors.
\newblock \emph{arXiv preprint arXiv:2402.03293}, 2024.

\bibitem[He et~al.(2024)He, Wang, Lin, Li, Ma, and Li]{10957983}
He, L., Wang, X., Lin, Y., Li, X., Ma, Y., and Li, Z.
\newblock Boann: Bayesian-optimized attentive neural network for classification.
\newblock In \emph{2024 International Conference on Image Processing, Computer Vision and Machine Learning (ICICML)}, pp.\  1860--1864, 2024.
\newblock \doi{10.1109/ICICML63543.2024.10957983}.

\bibitem[Jiang et~al.(2023)Jiang, Sablayrolles, Mensch, Bamford, Chaplot, Casas, Bressand, Lengyel, Lample, Saulnier, et~al.]{jiang2023mistral}
Jiang, A.~Q., Sablayrolles, A., Mensch, A., Bamford, C., Chaplot, D.~S., Casas, D. d.~l., Bressand, F., Lengyel, G., Lample, G., Saulnier, L., et~al.
\newblock Mistral 7b.
\newblock \emph{arXiv preprint arXiv:2310.06825}, 2023.

\bibitem[Jiang et~al.(2024)Jiang, Chen, Nguyen, Mervak, and Tan]{jiang2024gpt}
Jiang, Y., Chen, C., Nguyen, D., Mervak, B.~M., and Tan, C.
\newblock Gpt-4v cannot generate radiology reports yet.
\newblock \emph{arXiv preprint arXiv:2407.12176}, 2024.

\bibitem[Johnson et~al.(2016)Johnson, Pollard, Shen, wei H.~Lehman, Feng, Ghassemi, Moody, Szolovits, Celi, and Mark]{Johnson2016MIMICIIIAF}
Johnson, A. E.~W., Pollard, T.~J., Shen, L., wei H.~Lehman, L., Feng, M., Ghassemi, M.~M., Moody, B., Szolovits, P., Celi, L.~A., and Mark, R.~G.
\newblock Mimic-iii, a freely accessible critical care database.
\newblock \emph{Scientific Data}, 3, 2016.
\newblock URL \url{https://api.semanticscholar.org/CorpusID:33285731}.

\bibitem[Kry{\'s}ci{\'n}ski et~al.(2019)Kry{\'s}ci{\'n}ski, McCann, Xiong, and Socher]{kryscinski2019evaluating}
Kry{\'s}ci{\'n}ski, W., McCann, B., Xiong, C., and Socher, R.
\newblock Evaluating the factual consistency of abstractive text summarization.
\newblock \emph{arXiv preprint arXiv:1910.12840}, 2019.

\bibitem[Lee et~al.(2020)Lee, Yoon, Kim, Kim, Kim, So, and Kang]{lee2020biobert}
Lee, J., Yoon, W., Kim, S., Kim, D., Kim, S., So, C.~H., and Kang, J.
\newblock Biobert: a pre-trained biomedical language representation model for biomedical text mining.
\newblock \emph{Bioinformatics}, 36\penalty0 (4):\penalty0 1234--1240, 2020.

\bibitem[Leong \& Wu(2024)Leong and Wu]{leong2024nextgen}
Leong, H.~Y. and Wu, Y.
\newblock Why should next-gen llm multi-agent systems move beyond fixed architectures to dynamic, input-driven graphs?
\newblock SSRN Working Paper, 2024.
\newblock URL \url{https://ssrn.com/abstract=5276004}.

\bibitem[Leong et~al.(2024{\natexlab{a}})Leong, Gao, and Ji]{leong2024genai}
Leong, H.~Y., Gao, Y., and Ji, S.
\newblock A gen ai framework for medical note generation.
\newblock In \emph{Proceedings of the 2024 6th International Conference on Artificial Intelligence and Computer Applications (ICAICA)}, pp.\  423--429, 2024{\natexlab{a}}.
\newblock \doi{10.1109/ICAICA63239.2024.10823004}.

\bibitem[Leong et~al.(2024{\natexlab{b}})Leong, Gao, Ji, Zhang, and Pamuksuz]{leong2024efficient}
Leong, H.~Y., Gao, Y., Ji, S., Zhang, Y., and Pamuksuz, U.
\newblock Efficient fine-tuning of large language models for automated medical documentation.
\newblock In \emph{Proceedings of the 2024 4th International Conference on Digital Society and Intelligent Systems (DSInS)}, pp.\  204--209, Sydney, Australia, 2024{\natexlab{b}}.
\newblock \doi{10.1109/DSInS64146.2024.10992195}.

\bibitem[Li et~al.(2025{\natexlab{a}})Li, Jia, Wang, An, Li, Hwang, and Belongie]{li2025chatmotion}
Li, L., Jia, S., Wang, J., An, Z., Li, J., Hwang, J.-N., and Belongie, S.
\newblock Chatmotion: A multimodal multi-agent for human motion analysis.
\newblock \emph{arXiv preprint arXiv:2502.18180}, 2025{\natexlab{a}}.

\bibitem[Li et~al.(2025{\natexlab{b}})Li, Jia, Wang, Jiang, Zhou, Dai, Zhang, Wu, and Hwang]{li2025human}
Li, L., Jia, S., Wang, J., Jiang, Z., Zhou, F., Dai, J., Zhang, T., Wu, Z., and Hwang, J.-N.
\newblock Human motion instruction tuning.
\newblock In \emph{Proceedings of the Computer Vision and Pattern Recognition Conference}, pp.\  17582--17591, 2025{\natexlab{b}}.

\bibitem[Li et~al.(2024{\natexlab{a}})Li, Ma, Huang, Wang, Lin, and Zhang]{10800533}
Li, X., Ma, Y., Huang, Y., Wang, X., Lin, Y., and Zhang, C.
\newblock Synergized data efficiency and compression (sec) optimization for large language models.
\newblock In \emph{2024 4th International Conference on Electronic Information Engineering and Computer Science (EIECS)}, pp.\  586--591, 2024{\natexlab{a}}.
\newblock \doi{10.1109/EIECS63941.2024.10800533}.

\bibitem[Li et~al.()Li, Yang, Zeng, Dong, An, Xu, Tian, and Wu]{li2025frequency}
Li, Y., Yang, C., Zeng, H., Dong, Z., An, Z., Xu, Y., Tian, Y., and Wu, H.
\newblock Frequency-aligned knowledge distillation for lightweight spatiotemporal forecasting.
\newblock \doi{10.48550/arXiv.2507.02939}.

\bibitem[Li(2024)]{li2024advances}
Li, Z.
\newblock Advances in deep reinforcement learning for computer vision applications.
\newblock \emph{Journal of Industrial Engineering and Applied Science}, 2024.

\bibitem[Li et~al.(2024{\natexlab{b}})Li, Liu, Zhou, and Ma]{li2024chain}
Li, Z., Liu, H., Zhou, D., and Ma, T.
\newblock Chain of thought empowers transformers to solve inherently serial problems.
\newblock \emph{arXiv preprint arXiv:2402.12875}, 2024{\natexlab{b}}.

\bibitem[Li et~al.(2025{\natexlab{c}})Li, Ji, Ling, and Liu]{11079788}
Li, Z., Ji, Q., Ling, X., and Liu, Q.
\newblock A comprehensive review of multi-agent reinforcement learning in video games.
\newblock \emph{IEEE Transactions on Games}, 2025{\natexlab{c}}.
\newblock \doi{10.1109/TG.2025.3588809}.

\bibitem[Liang et~al.(2024)Liang, He, Tao, Xia, Wang, Shi, Wang, and Yang]{liang2024cmat}
Liang, X., He, Y., Tao, M., Xia, Y., Wang, J., Shi, T., Wang, J., and Yang, J.
\newblock Cmat: A multi-agent collaboration tuning framework for enhancing small language models.
\newblock \emph{arXiv preprint arXiv:2404.01663}, 2024.

\bibitem[Liang et~al.(2025)Liang, Xiang, and Li]{liang2025search}
Liang, Y., Xiang, D., and Li, X.
\newblock Search: A self-evolving framework for network architecture optimization.
\newblock \emph{Neurocomputing}, pp.\  130980, 2025.

\bibitem[Liao et~al.(2024)Liao, Zhao, Chen, Li, Cremers, and Liu]{liao2024globalpointer}
Liao, B., Zhao, Z., Chen, L., Li, H., Cremers, D., and Liu, P.
\newblock Globalpointer: Large-scale plane adjustment with bi-convex relaxation.
\newblock In \emph{European Conference on Computer Vision}, pp.\  360--376. Springer, 2024.

\bibitem[Liao et~al.(2025)Liao, Zhao, Li, Zhou, Zeng, Li, and Liu]{liao2025convex}
Liao, B., Zhao, Z., Li, H., Zhou, Y., Zeng, Y., Li, H., and Liu, P.
\newblock Convex relaxation for robust vanishing point estimation in manhattan world.
\newblock In \emph{Proceedings of the Computer Vision and Pattern Recognition Conference}, pp.\  15823--15832, 2025.

\bibitem[Lin et~al.(2024)Lin, Zheng, Xue, Fu, Li, and Shen]{lin2024motion}
Lin, B., Zheng, J., Xue, C., Fu, L., Li, Y., and Shen, Q.
\newblock Motion-aware correlation filter-based object tracking in satellite videos.
\newblock \emph{IEEE Transactions on Geoscience and Remote Sensing}, 62:\penalty0 1--13, 2024.

\bibitem[Lin(2004)]{lin-2004-rouge}
Lin, C.-Y.
\newblock {ROUGE}: A package for automatic evaluation of summaries.
\newblock In \emph{Text Summarization Branches Out}, pp.\  74--81, Barcelona, Spain, July 2004. Association for Computational Linguistics.
\newblock URL \url{https://aclanthology.org/W04-1013/}.

\bibitem[Lin et~al.(2022)Lin, Shi, Xiang, Zeng, Gong, Liu, Liu, Chen, Xia, and Chen]{lin2022construction}
Lin, Y., Shi, M., Xiang, D., Zeng, P., Gong, Z., Liu, H., Liu, Q., Chen, Z., Xia, J., and Chen, Z.
\newblock Construction of an end-to-end regression neural network for the determination of a quantitative index sagittal root inclination.
\newblock \emph{Journal of Periodontology}, 93\penalty0 (12):\penalty0 1951--1960, 2022.

\bibitem[Liu et~al.()Liu, Zhang, Wang, Xiang, and Xie]{Liu04072025}
Liu, C., Zhang, Z., Wang, M., Xiang, S., and Xie, G.
\newblock A novel cross fusion model with fine-grained detail reconstruction for remote sensing image pan-sharpening.
\newblock \emph{Geo-spatial Information Science}.
\newblock \doi{10.1080/10095020.2024.2416899}.
\newblock URL \url{https://doi.org/10.1080/10095020.2024.2416899}.

\bibitem[Liu et~al.(2023)Liu, Wei, Zhang, Feng, and Xiang]{10294268}
Liu, C., Wei, L., Zhang, Z., Feng, X., and Xiang, S.
\newblock Recursive self-attention modules-based network for panchromatic and multispectral image fusion.
\newblock \emph{IEEE Journal of Selected Topics in Applied Earth Observations and Remote Sensing}, 2023.
\newblock \doi{10.1109/JSTARS.2023.3327167}.

\bibitem[Liu et~al.(2025{\natexlab{a}})Liu, Bao, and Zhang]{10813395}
Liu, C., Bao, L., and Zhang, Z.
\newblock A spatial–temporal difference aggregation network for gaofen-2 multitemporal image in cropland change area.
\newblock \emph{IEEE Journal of Selected Topics in Applied Earth Observations and Remote Sensing}, 2025{\natexlab{a}}.
\newblock \doi{10.1109/JSTARS.2024.3522066}.

\bibitem[Liu et~al.(2025{\natexlab{b}})Liu, Zhang, and Wang]{10934049}
Liu, C., Zhang, Z., and Wang, M.
\newblock Ramsf: A novel generic framework for optical remote sensing multimodal spatial-spectral fusion.
\newblock \emph{IEEE Transactions on Geoscience and Remote Sensing}, 2025{\natexlab{b}}.
\newblock \doi{10.1109/TGRS.2025.3552937}.

\bibitem[Liu et~al.(2024)Liu, Chen, Ji, Zhou, Chen, and Wang]{liu2024rag}
Liu, W., Chen, J., Ji, K., Zhou, L., Chen, W., and Wang, B.
\newblock Rag-instruct: Boosting llms with diverse retrieval-augmented instructions.
\newblock \emph{arXiv preprint arXiv:2501.00353}, 2024.

\bibitem[Liu et~al.(2025{\natexlab{c}})Liu, Xu, Yu, Lin, Ji, Chen, Xu, Wang, Shang, and Wang]{liu2025qfft}
Liu, W., Xu, J., Yu, F., Lin, Y., Ji, K., Chen, W., Xu, Y., Wang, Y., Shang, L., and Wang, B.
\newblock Qfft, question-free fine-tuning for adaptive reasoning.
\newblock \emph{arXiv preprint arXiv:2506.12860}, 2025{\natexlab{c}}.

\bibitem[Liu et~al.(2025{\natexlab{d}})Liu, Lu, Wang, and Wu]{liu2025amsf}
Liu, X., Lu, Y., Wang, X., and Wu, X.
\newblock Training-free multi-style fusion through reference-based adaptive modulation.
\newblock \emph{arXiv preprint arXiv:2509.18602}, 2025{\natexlab{d}}.
\newblock \doi{10.48550/arXiv.2509.18602}.
\newblock URL \url{https://arxiv.org/abs/2509.18602}.
\newblock Accepted at ACPR 2025 (oral).

\bibitem[Loeschcke et~al.(2024)Loeschcke, Toftrup, Kastoryano, Belongie, and Sn{\ae}bjarnarson]{loeschcke2024loqt}
Loeschcke, S., Toftrup, M., Kastoryano, M.~J., Belongie, S., and Sn{\ae}bjarnarson, V.
\newblock Loqt: Low rank adapters for quantized training.
\newblock \emph{arXiv preprint arXiv:2405.16528}, 2024.

\bibitem[Lu et~al.(2022{\natexlab{a}})Lu, Wu, Parker, Cheng, Saylor, Van~Eyk, Yu, Clarke, Herrington, and Wang]{lu2022cot}
Lu, Y., Wu, C.-T., Parker, S.~J., Cheng, Z., Saylor, G., Van~Eyk, J.~E., Yu, G., Clarke, R., Herrington, D.~M., and Wang, Y.
\newblock Cot: an efficient and accurate method for detecting marker genes among many subtypes.
\newblock \emph{Bioinformatics Advances}, pp.\  vbac037, 2022{\natexlab{a}}.

\bibitem[Lu et~al.(2022{\natexlab{b}})Lu, Yang, Zhang, Chen, Chen, Xuan, Wang, and Yang]{lu2022understanding}
Lu, Y., Yang, W., Zhang, Y., Chen, Z., Chen, J., Xuan, Q., Wang, Z., and Yang, X.
\newblock Understanding the dynamics of dnns using graph modularity.
\newblock In \emph{European Conference on Computer Vision}, pp.\  225--242. Springer, 2022{\natexlab{b}}.

\bibitem[Lu et~al.(2023{\natexlab{a}})Lu, Chen, Zhang, Gu, Zhang, Zhang, Yang, Xuan, Wang, and You]{lu2023can}
Lu, Y., Chen, X., Zhang, Y., Gu, J., Zhang, T., Zhang, Y., Yang, X., Xuan, Q., Wang, K., and You, Y.
\newblock Can pre-trained models assist in dataset distillation?
\newblock \emph{arXiv preprint arXiv:2310.03295}, 2023{\natexlab{a}}.

\bibitem[Lu et~al.(2023{\natexlab{b}})Lu, Sato, and Wang]{lu2023deep}
Lu, Y., Sato, K., and Wang, J.
\newblock Deep learning based multi-label image classification of protest activities.
\newblock \emph{arXiv preprint arXiv:2301.04212}, 2023{\natexlab{b}}.

\bibitem[Lu et~al.(2024{\natexlab{a}})Lu, Cheng, Fang, Wang, Wei, Xu, Xuan, Yang, and Zhu]{lu2024reassessing}
Lu, Y., Cheng, H., Fang, Y., Wang, Z., Wei, J., Xu, D., Xuan, Q., Yang, X., and Zhu, Z.
\newblock Reassessing layer pruning in llms: New insights and methods.
\newblock \emph{arXiv preprint arXiv:2411.15558}, 2024{\natexlab{a}}.

\bibitem[Lu et~al.(2024{\natexlab{b}})Lu, Zhu, Li, Xu, Lin, Xuan, and Yang]{lu2024generic}
Lu, Y., Zhu, Y., Li, Y., Xu, D., Lin, Y., Xuan, Q., and Yang, X.
\newblock A generic layer pruning method for signal modulation recognition deep learning models.
\newblock \emph{IEEE Transactions on Cognitive Communications and Networking}, 2024{\natexlab{b}}.

\bibitem[Nguyen et~al.(2024)Nguyen, Hains, Aziz-Zanjani, Dalsass, Farooqee, Lu, Jackson, and Van~Rechem]{nguyen2024absence}
Nguyen, L.~T., Hains, A.~E., Aziz-Zanjani, M.~O., Dalsass, M., Farooqee, S.~B., Lu, Y., Jackson, P.~K., and Van~Rechem, C.
\newblock Absence of smarcb1 in rhabdoid tumor cells increases sensitivity to translation inhibition and alters translation efficiency of specific mrnas.
\newblock \emph{Journal of Biological Chemistry}, 300\penalty0 (12), 2024.

\bibitem[Papineni et~al.(2002)Papineni, Roukos, Ward, and Zhu]{Papineni2002BleuAM}
Papineni, K., Roukos, S., Ward, T., and Zhu, W.-J.
\newblock Bleu: a method for automatic evaluation of machine translation.
\newblock In \emph{Annual Meeting of the Association for Computational Linguistics}, 2002.
\newblock URL \url{https://api.semanticscholar.org/CorpusID:11080756}.

\bibitem[Parker et~al.(2024)Parker, Mao, Wang, Seals, Kim, Lu, bhardwaj, Du, Karere, Caudell, et~al.]{parker2024mapping}
Parker, S., Mao, C., Wang, Y., Seals, A., Kim, D.-K., Lu, Y., bhardwaj, s., Du, D., Karere, G., Caudell, D., et~al.
\newblock Mapping the molecular progression of human coronary atherosclerosis confirms key role of smooth muscle cell phenotype and highlights novel regulators of phenotypic fate.
\newblock \emph{Circulation}, 150\penalty0 (Suppl\_1):\penalty0 A4139230--A4139230, 2024.

\bibitem[Peng et~al.(2019)Peng, Yan, and Lu]{peng-etal-2019-transfer}
Peng, Y., Yan, S., and Lu, Z.
\newblock Transfer learning in biomedical natural language processing: An evaluation of {BERT} and {ELM}o on ten benchmarking datasets.
\newblock In Demner-Fushman, D., Cohen, K.~B., Ananiadou, S., and Tsujii, J. (eds.), \emph{Proceedings of the 18th BioNLP Workshop and Shared Task}, pp.\  58--65, Florence, Italy, August 2019. Association for Computational Linguistics.
\newblock \doi{10.18653/v1/W19-5006}.
\newblock URL \url{https://aclanthology.org/W19-5006/}.

\bibitem[Peng et~al.(2020)Peng, Chen, and Lu]{peng2020empirical}
Peng, Y., Chen, Q., and Lu, Z.
\newblock An empirical study of multi-task learning on bert for biomedical text mining.
\newblock \emph{arXiv preprint arXiv:2005.02799}, 2020.

\bibitem[Qinsi et~al.()Qinsi, Ke, Tomizuka, Keutzer, and Xu]{qinsidobi}
Qinsi, W., Ke, J., Tomizuka, M., Keutzer, K., and Xu, C.
\newblock Dobi-svd: Differentiable svd for llm compression and some new perspectives.
\newblock In \emph{The Thirteenth International Conference on Learning Representations}.

\bibitem[Qiu et~al.(2025)Qiu, Gao, Li, Leong, and Zhang]{Qiu2025}
Qiu, T., Gao, J., Li, J., Leong, H., and Zhang, L.
\newblock Intentvcnet: Bridging spatio-temporal gaps for intention-oriented controllable video captioning.
\newblock In \emph{Proceedings of ACM Multimedia (MM) 2025}, 2025.
\newblock \doi{10.2139/ssrn.5374737}.
\newblock URL \url{https://ssrn.com/abstract=5374737}.
\newblock Accepted, In Press.

\bibitem[Romanov \& Shivade(2018)Romanov and Shivade]{romanov2018lessons}
Romanov, A. and Shivade, C.
\newblock Lessons from natural language inference in the clinical domain.
\newblock \emph{arXiv preprint arXiv:1808.06752}, 2018.

\bibitem[Rust et~al.(2022)Rust, Lotz, Bugliarello, Salesky, de~Lhoneux, and Elliott]{rust2022language}
Rust, P., Lotz, J.~F., Bugliarello, E., Salesky, E., de~Lhoneux, M., and Elliott, D.
\newblock Language modelling with pixels.
\newblock \emph{arXiv preprint arXiv:2207.06991}, 2022.

\bibitem[Shi et~al.(2025)Shi, Ma, Liu, Zhao, Hwang, and Li]{shi2025explaining}
Shi, J., Ma, Q., Liu, H., Zhao, H., Hwang, J.-N., and Li, L.
\newblock Explaining context length scaling and bounds for language models.
\newblock \emph{arXiv preprint arXiv:2502.01481}, 2025.

\bibitem[Shi et~al.(2024)Shi, Gong, Zeng, Xiang, Cai, Liu, Chen, Liu, Chen, Zhang, et~al.]{shi2024multi}
Shi, M., Gong, Z., Zeng, P., Xiang, D., Cai, G., Liu, H., Chen, S., Liu, R., Chen, Z., Zhang, X., et~al.
\newblock Multi-quantifying maxillofacial traits via a demographic parity-based ai model.
\newblock \emph{BME frontiers}, 5:\penalty0 0054, 2024.

\bibitem[Sun et~al.(2013)Sun, Rumshisky, and Uzuner]{sun2013evaluating}
Sun, W., Rumshisky, A., and Uzuner, O.
\newblock Evaluating temporal relations in clinical text: 2012 i2b2 challenge.
\newblock \emph{Journal of the American Medical Informatics Association}, 20\penalty0 (5):\penalty0 806--813, 2013.

\bibitem[Sunmola et~al.(2025)Sunmola, Zhao, Schmidgall, Wang, Scheikl, and Krieger]{sunmola2025surgical}
Sunmola, I.~O., Zhao, Z., Schmidgall, S., Wang, Y., Scheikl, P.~M., and Krieger, A.
\newblock Surgical gaussian surfels: Highly accurate real-time surgical scene rendering.
\newblock \emph{arXiv preprint arXiv:2503.04079}, 2025.

\bibitem[Team et~al.(2024)Team, Mesnard, Hardin, Dadashi, Bhupatiraju, Pathak, Sifre, Rivi{\`e}re, Kale, Love, et~al.]{team2024gemma}
Team, G., Mesnard, T., Hardin, C., Dadashi, R., Bhupatiraju, S., Pathak, S., Sifre, L., Rivi{\`e}re, M., Kale, M.~S., Love, J., et~al.
\newblock Gemma: Open models based on gemini research and technology.
\newblock \emph{arXiv preprint arXiv:2403.08295}, 2024.

\bibitem[Uzuner et~al.(2011)Uzuner, South, Shen, and DuVall]{uzuner20112010}
Uzuner, {\"O}., South, B.~R., Shen, S., and DuVall, S.~L.
\newblock 2010 i2b2/va challenge on concepts, assertions, and relations in clinical text.
\newblock \emph{Journal of the American Medical Informatics Association}, 18\penalty0 (5):\penalty0 552--556, 2011.

\bibitem[Wang et~al.(2025{\natexlab{a}})Wang, Li, Zhou, Leong, Zhao, Ye, Deng, Luo, and Vasconcelos]{Wang2025e}
Wang, B., Li, Y., Zhou, Q., Leong, H., Zhao, T., Ye, L., Deng, H., Luo, D., and Vasconcelos, N.
\newblock Do vision language models infer human intention without visual perspective-taking? towards a scalable ``one-image-probe-all'' dataset.
\newblock In \emph{Proceedings of the ICML 2025 Workshop on Assessing World Models}, 2025{\natexlab{a}}.
\newblock URL \url{https://openreview.net/forum?id=iekoq1rv80}.

\bibitem[Wang et~al.(2024{\natexlab{a}})Wang, Zhang, He, Zhang, Song, Shi, Li, Xu, Wu, Yi, et~al.]{wang2024enhancing}
Wang, J., Zhang, Z., He, Y., Zhang, Z., Song, Y., Shi, T., Li, Y., Xu, H., Wu, K., Yi, X., et~al.
\newblock Enhancing code llms with reinforcement learning in code generation: A survey.
\newblock \emph{arXiv preprint arXiv:2412.20367}, 2024{\natexlab{a}}.

\bibitem[Wang et~al.(2023{\natexlab{a}})Wang, Xu, Lan, Hu, Lan, Lee, and Lim]{wang2023plan}
Wang, L., Xu, W., Lan, Y., Hu, Z., Lan, Y., Lee, R. K.-W., and Lim, E.-P.
\newblock Plan-and-solve prompting: Improving zero-shot chain-of-thought reasoning by large language models.
\newblock \emph{arXiv preprint arXiv:2305.04091}, 2023{\natexlab{a}}.

\bibitem[Wang et~al.(2023{\natexlab{b}})Wang, Ke, Liang, and Zhang]{wang2023mathnas}
Wang, Q., Ke, J., Liang, Z., and Zhang, S.
\newblock Mathnas: if blocks have a role in mathematical architecture design.
\newblock \emph{Advances in Neural Information Processing Systems}, 36:\penalty0 47475--47486, 2023{\natexlab{b}}.

\bibitem[Wang et~al.(2024{\natexlab{b}})Wang, Vahidian, Ye, Gu, Zhang, and Chen]{wang2024coreinfer}
Wang, Q., Vahidian, S., Ye, H., Gu, J., Zhang, J., and Chen, Y.
\newblock Coreinfer: Accelerating large language model inference with semantics-inspired adaptive sparse activation.
\newblock \emph{arXiv preprint arXiv:2410.18311}, 2024{\natexlab{b}}.

\bibitem[Wang et~al.(2025{\natexlab{b}})Wang, Ke, Ye, Lin, Fu, Zhang, Keutzer, Xu, and Chen]{wang2025anglesdontlieunlocking}
Wang, Q., Ke, J., Ye, H., Lin, Y., Fu, Y., Zhang, J., Keutzer, K., Xu, C., and Chen, Y.
\newblock Angles don't lie: Unlocking training-efficient rl through the model's own signals, 2025{\natexlab{b}}.
\newblock URL \url{https://arxiv.org/abs/2506.02281}.

\bibitem[Wang et~al.(2025{\natexlab{c}})Wang, Ye, Chung, Liu, Lin, Kuo, Ma, Zhang, and Chen]{wang2025corematching}
Wang, Q., Ye, H., Chung, M.-Y., Liu, Y., Lin, Y., Kuo, M., Ma, M., Zhang, J., and Chen, Y.
\newblock Corematching: A co-adaptive sparse inference framework with token and neuron pruning for comprehensive acceleration of vision-language models.
\newblock \emph{arXiv preprint arXiv:2505.19235}, 2025{\natexlab{c}}.

\bibitem[Wang et~al.(2024{\natexlab{c}})Wang, Zhang, and Dodgson]{wang2024scantd}
Wang, Y., Zhang, F.-L., and Dodgson, N.~A.
\newblock Scantd: 360° scanpath prediction based on time-series diffusion.
\newblock In \emph{Proceedings of the 32nd ACM International Conference on Multimedia}, 2024{\natexlab{c}}.
\newblock \doi{10.1145/3664647.3681315}.

\bibitem[Wang et~al.(2025{\natexlab{d}})Wang, Wang, Zhong, Zhu, and Li]{wang2025applications}
Wang, Y., Wang, Z., Zhong, J., Zhu, D., and Li, W.
\newblock Applications of small language models in medical imaging classification with a focus on prompt strategies.
\newblock \emph{arXiv preprint arXiv:2508.13378}, 2025{\natexlab{d}}.

\bibitem[Wang et~al.(2025{\natexlab{e}})Wang, Zhang, and Dodgson]{Wang2025Scanpath360}
Wang, Y., Zhang, F.-L., and Dodgson, N.~A.
\newblock Target scanpath-guided 360-degree image enhancement.
\newblock In \emph{Proceedings of the 39th AAAI Conference on Artificial Intelligence (AAAI)}. AAAI Press, 2025{\natexlab{e}}.
\newblock \doi{10.1609/aaai.v39i8.32881}.
\newblock URL \url{https://doi.org/10.1609/aaai.v39i8.32881}.

\bibitem[Wang et~al.(2025{\natexlab{f}})Wang, Zhong, and Kumar]{wang2025systematic}
Wang, Y., Zhong, J., and Kumar, R.
\newblock A systematic review of machine learning applications in infectious disease prediction, diagnosis, and outbreak forecasting.
\newblock 2025{\natexlab{f}}.

\bibitem[Wu et~al.(2025)Wu, Liu, Zhao, and Wu]{wu2025pdls}
Wu, Y., Liu, X., Zhao, C., and Wu, X.
\newblock Prompt-guided dual latent steering for inversion problems.
\newblock \emph{arXiv preprint arXiv:2509.18619}, 2025.
\newblock \doi{10.48550/arXiv.2509.18619}.
\newblock URL \url{https://arxiv.org/abs/2509.18619}.
\newblock Accepted at DICTA 2025 (oral).

\bibitem[Xiang et~al.(2025{\natexlab{a}})Xiang, Sun, Qi, Zhao, and Qi]{ding2025artificial}
Xiang, D., Sun, T., Qi, Y., Zhao, Z., and Qi, Y.
\newblock Artificial intelligence applications in power electronics.
\newblock 2025{\natexlab{a}}.

\bibitem[Xiang et~al.(2025{\natexlab{b}})Xiang, Xu, Chu, Shen, Ding, and Zhang]{xiang2025promptsculptor}
Xiang, D., Xu, W., Chu, K., Shen, Z., Ding, T., and Zhang, W.
\newblock Promptsculptor: Multi-agent based text-to-image prompt optimization.
\newblock \emph{arXiv preprint arXiv:2509.12446}, 2025{\natexlab{b}}.

\bibitem[Xu et~al.(2023)Xu, Xie, Qin, Tao, and Wang]{xu2023parameter}
Xu, L., Xie, H., Qin, S.-Z.~J., Tao, X., and Wang, F.~L.
\newblock Parameter-efficient fine-tuning methods for pretrained language models: A critical review and assessment.
\newblock \emph{arXiv preprint arXiv:2312.12148}, 2023.

\bibitem[Xu et~al.(2025{\natexlab{a}})Xu, Xiang, Liu, Wang, Ma, Zhang, Xu, and Zhang]{xu2025finmultitime}
Xu, W., Xiang, D., Liu, Y., Wang, X., Ma, Y., Zhang, L., Xu, C., and Zhang, J.
\newblock Finmultitime: A four-modal bilingual dataset for financial time-series analysis.
\newblock \emph{arXiv preprint arXiv:2506.05019}, 2025{\natexlab{a}}.

\bibitem[Xu et~al.(2025{\natexlab{b}})Xu, Xiang, Wang, Hu, Zhang, Chen, and Lu]{xu2025learning}
Xu, W., Xiang, D., Wang, R., Hu, Y., Zhang, L., Chen, J., and Lu, Z.
\newblock Learning explainable stock predictions with tweets using mixture of experts.
\newblock \emph{arXiv preprint arXiv:2507.20535}, 2025{\natexlab{b}}.

\bibitem[Xue et~al.(2024)Xue, Zhong, Liang, Xia, and Song]{xue2024unifying}
Xue, C., Zhong, B., Liang, Q., Xia, H., and Song, S.
\newblock Unifying motion and appearance cues for visual tracking via shared queries.
\newblock \emph{IEEE Transactions on Circuits and Systems for Video Technology}, 2024.

\bibitem[Xue et~al.(2025)Xue, Zhong, Liang, Zheng, Li, Xue, and Song]{xue2025similarity}
Xue, C., Zhong, B., Liang, Q., Zheng, Y., Li, N., Xue, Y., and Song, S.
\newblock Similarity-guided layer-adaptive vision transformer for uav tracking.
\newblock In \emph{Proceedings of the Computer Vision and Pattern Recognition Conference}, pp.\  6730--6740, 2025.

\bibitem[Yan et~al.(2022)Yan, McAuley, Lu, Du, Chang, Gentili, and Hsu]{yan2022radbert}
Yan, A., McAuley, J., Lu, X., Du, J., Chang, E.~Y., Gentili, A., and Hsu, C.-N.
\newblock Radbert: adapting transformer-based language models to radiology.
\newblock \emph{Radiology: Artificial Intelligence}, 4\penalty0 (4):\penalty0 e210258, 2022.

\bibitem[Yao et~al.(2025)Yao, Cheng, Huang, and Li]{yao2025countllm}
Yao, Z., Cheng, X., Huang, Z., and Li, L.
\newblock Countllm: Towards generalizable repetitive action counting via large language model.
\newblock In \emph{Proceedings of the Computer Vision and Pattern Recognition Conference}, pp.\  19143--19153, 2025.

\bibitem[Yi et~al.(2025)Yi, He, Wang, Song, Qian, Yuan, Sun, Xin, Tang, Li, et~al.]{yi2025score}
Yi, Q., He, Y., Wang, J., Song, X., Qian, S., Yuan, X., Sun, L., Xin, Y., Tang, J., Li, K., et~al.
\newblock Score: Story coherence and retrieval enhancement for ai narratives.
\newblock \emph{arXiv preprint arXiv:2503.23512}, 2025.

\bibitem[Yuan et~al.(2024)Yuan, Huang, Ma, Li, Li, Shi, and Zhou]{10.1117/12.3049486}
Yuan, Y., Huang, Y., Ma, Y., Li, X., Li, Z., Shi, Y., and Zhou, H.
\newblock {Rhyme-aware Chinese lyric generator based on GPT}.
\newblock In Zhang, W. and Lu, Q. (eds.), \emph{Fourth International Conference on Advanced Algorithms and Neural Networks (AANN 2024)}, volume 13416, pp.\  134162P. International Society for Optics and Photonics, SPIE, 2024.
\newblock \doi{10.1117/12.3049486}.
\newblock URL \url{https://doi.org/10.1117/12.3049486}.

\bibitem[Zeng et~al.(2025)Zeng, Chang, Xie, Liu, Bai, Pan, Xu, and Wei]{zeng2025FSDrive}
Zeng, S., Chang, X., Xie, M., Liu, X., Bai, Y., Pan, Z., Xu, M., and Wei, X.
\newblock Futuresightdrive: Thinking visually with spatio-temporal cot for autonomous driving.
\newblock \emph{arXiv preprint arXiv:2505.17685}, 2025.

\bibitem[Zhang et~al.(2025{\natexlab{a}})Zhang, Huang, Li, Xiao, Leong, Zhang, Long, Wang, and Xu]{Zhang2025}
Zhang, H., Huang, B., Li, Z., Xiao, X., Leong, H., Zhang, Z., Long, X., Wang, T., and Xu, H.
\newblock Sensitivity-lora: Low-load sensitivity-based fine-tuning for large language models.
\newblock In \emph{Findings of the 2025 Conference on Empirical Methods in Natural Language Processing (EMNLP)}, 2025{\natexlab{a}}.
\newblock URL \url{https://openreview.net/forum?id=5erJwHj6CC#discussion}.

\bibitem[Zhang et~al.(2025{\natexlab{b}})Zhang, Gao, Ouyang, Zhu, and Leong]{Zhang2025t}
Zhang, J., Gao, J., Ouyang, W., Zhu, W., and Leong, H.
\newblock Time-llama: Adapting large language models for time series modeling via dynamic low-rank adaptation.
\newblock In \emph{Proceedings of the 63rd Annual Meeting of the Association for Computational Linguistics (Volume 4: Student Research Workshop)}. Association for Computational Linguistics (ACL 2025), 2025{\natexlab{b}}.
\newblock ISBN 979-8-89176-254-1.
\newblock \doi{10.18653/v1/2025.acl-srw.90}.
\newblock URL \url{https://aclanthology.org/2025.acl-srw.90/}.
\newblock Poster.

\bibitem[Zhang et~al.(2023)Zhang, Wu, Mo, Nie, and Agrawal]{zhang2023moqagpt}
Zhang, L., Wu, Y., Mo, F., Nie, J.-Y., and Agrawal, A.
\newblock Moqagpt: Zero-shot multi-modal open-domain question answering with large language model.
\newblock \emph{arXiv preprint arXiv:2310.13265}, 2023.

\bibitem[Zhang et~al.(2020)Zhang, Kishore, Wu, Weinberger, and Artzi]{zhang2020bertscoreevaluatingtextgeneration}
Zhang, T., Kishore, V., Wu, F., Weinberger, K.~Q., and Artzi, Y.
\newblock Bertscore: Evaluating text generation with bert, 2020.
\newblock URL \url{https://arxiv.org/abs/1904.09675}.

\bibitem[Zhang et~al.(2024{\natexlab{a}})Zhang, Li, Tian, and Liu]{11117915}
Zhang, W., Li, Z., Tian, Y., and Liu, L.
\newblock Research on temperature prediction based on rf-lstm modeling.
\newblock In \emph{2024 6th International Academic Exchange Conference on Science and Technology Innovation (IAECST)}, 2024{\natexlab{a}}.
\newblock \doi{10.1109/IAECST64597.2024.11117915}.

\bibitem[Zhang(2019)]{zhang2019evaluating}
Zhang, Y.
\newblock Evaluating the factual correctness for abstractive summarization.
\newblock \emph{CS230 Project}, 2019.

\bibitem[Zhang et~al.(2025{\natexlab{c}})Zhang, Liu, Tao, Chen, Fei, Che, and Qin]{zhang2025vitcot}
Zhang, Y., Liu, X., Tao, R., Chen, Q., Fei, H., Che, W., and Qin, L.
\newblock Vitcot: Video-text interleaved chain-of-thought for boosting video understanding in large language models.
\newblock \emph{arXiv preprint arXiv:2507.09876}, 2025{\natexlab{c}}.

\bibitem[Zhang et~al.(2025{\natexlab{d}})Zhang, Liu, Zhou, Chen, Fei, Lu, and Qin]{zhang2025cchall}
Zhang, Y., Liu, X., Zhou, R., Chen, Q., Fei, H., Lu, W., and Qin, L.
\newblock Cchall: A novel benchmark for joint cross-lingual and cross-modal hallucinations detection in large language models.
\newblock \emph{arXiv preprint arXiv:2505.19108}, 2025{\natexlab{d}}.

\bibitem[Zhang et~al.()Zhang, Wang, Li, Wang, and Zheng]{sym17071087}
Zhang, Z., Wang, J., Li, Z., Wang, Y., and Zheng, J.
\newblock Anncoder: A mti-agent-based code generation and optimization model.
\newblock \emph{Symmetry}.
\newblock ISSN 2073-8994.
\newblock \doi{10.3390/sym17071087}.
\newblock URL \url{https://www.mdpi.com/2073-8994/17/7/1087}.

\bibitem[Zhang et~al.(2024{\natexlab{b}})Zhang, Liu, Wei, and Xiang]{10741347}
Zhang, Z., Liu, C., Wei, L., and Xiang, S.
\newblock Mmapp: Multibranch and multiscale adaptive progressive pyramid network for multispectral image pansharpening.
\newblock \emph{IEEE Journal of Selected Topics in Applied Earth Observations and Remote Sensing}, 2024{\natexlab{b}}.
\newblock \doi{10.1109/JSTARS.2024.3490755}.

\bibitem[Zhao(2024)]{zhao2024balf}
Zhao, Z.
\newblock Balf: Simple and efficient blur aware local feature detector.
\newblock In \emph{Proceedings of the IEEE/CVF Winter Conference on Applications of Computer Vision}, pp.\  3362--3372, 2024.

\bibitem[Zhao \& Chen(2023)Zhao and Chen]{zhao2023benchmark}
Zhao, Z. and Chen, B.~M.
\newblock Benchmark for evaluating initialization of visual-inertial odometry.
\newblock In \emph{2023 42nd Chinese Control Conference (CCC)}, pp.\  3935--3940. IEEE, 2023.

\bibitem[Zheng et~al.(2022)Zheng, Zhong, Liang, Tang, Ji, and Li]{zheng2022leveraging}
Zheng, Y., Zhong, B., Liang, Q., Tang, Z., Ji, R., and Li, X.
\newblock Leveraging local and global cues for visual tracking via parallel interaction network.
\newblock \emph{IEEE Transactions on Circuits and Systems for Video Technology}, 2022.

\bibitem[Zheng et~al.(2023)Zheng, Zhong, Liang, Li, Ji, and Li]{zheng2023toward}
Zheng, Y., Zhong, B., Liang, Q., Li, G., Ji, R., and Li, X.
\newblock Toward unified token learning for vision-language tracking.
\newblock \emph{IEEE Transactions on Circuits and Systems for Video Technology}, 2023.

\bibitem[Zheng et~al.(2024)Zheng, Zhong, Liang, Mo, Zhang, and Li]{zheng2024odtrack}
Zheng, Y., Zhong, B., Liang, Q., Mo, Z., Zhang, S., and Li, X.
\newblock Odtrack: Online dense temporal token learning for visual tracking.
\newblock In \emph{Proceedings of the AAAI conference on artificial intelligence}, 2024.

\bibitem[Zheng et~al.(2025{\natexlab{a}})Zheng, Zhong, Liang, Li, and Song]{zheng2025decoupled}
Zheng, Y., Zhong, B., Liang, Q., Li, N., and Song, S.
\newblock Decoupled spatio-temporal consistency learning for self-supervised tracking.
\newblock In \emph{Proceedings of the AAAI Conference on Artificial Intelligence}, 2025{\natexlab{a}}.

\bibitem[Zheng et~al.(2025{\natexlab{b}})Zheng, Zhong, Liang, Zhang, Li, Li, and Ji]{zheng2025towards}
Zheng, Y., Zhong, B., Liang, Q., Zhang, S., Li, G., Li, X., and Ji, R.
\newblock Towards universal modal tracking with online dense temporal token learning.
\newblock \emph{IEEE Transactions on Pattern Analysis and Machine Intelligence}, 2025{\natexlab{b}}.

\bibitem[Zhou et~al.(2025{\natexlab{a}})Zhou, Jiang, Luan, Meng, Wang, Dong, Zhou, and He]{10969987}
Zhou, C., Jiang, R., Luan, F., Meng, S., Wang, Z., Dong, Y., Zhou, Y., and He, B.
\newblock Dual-arm robotic fabric manipulation with quasi-static and dynamic primitives for rapid garment flattening.
\newblock \emph{IEEE/ASME Transactions on Mechatronics}, 2025{\natexlab{a}}.
\newblock \doi{10.1109/TMECH.2025.3556283}.

\bibitem[Zhou et~al.(2025{\natexlab{b}})Zhou, Luan, hujiarui, Meng, Wang, Dong, Zhou, and He]{changshi2025learning}
Zhou, C., Luan, F., hujiarui, Meng, S., Wang, Z., Dong, Y., Zhou, Y., and He, B.
\newblock Learning efficient robotic garment manipulation with standardization.
\newblock In \emph{Forty-second International Conference on Machine Learning}, 2025{\natexlab{b}}.
\newblock URL \url{https://openreview.net/forum?id=pT4gJ0PgGD}.

\bibitem[Zhou et~al.(2025{\natexlab{c}})Zhou, Xu, Gu, Wang, Cheng, Zhang, Dong, Hayashibe, Zhou, and He]{zhou2025languageguidedlonghorizonmanipulation}
Zhou, C., Xu, H., Gu, N., Wang, Z., Cheng, B., Zhang, P., Dong, Y., Hayashibe, M., Zhou, Y., and He, B.
\newblock Language-guided long horizon manipulation with llm-based planning and visual perception, 2025{\natexlab{c}}.
\newblock URL \url{https://arxiv.org/abs/2509.02324}.

\bibitem[Zhou et~al.(2025{\natexlab{d}})Zhou, Xu, Hu, Luan, Wang, Dong, Zhou, and He]{10966003}
Zhou, C., Xu, H., Hu, J., Luan, F., Wang, Z., Dong, Y., Zhou, Y., and He, B.
\newblock Ssfold: Learning to fold arbitrary crumpled cloth using graph dynamics from human demonstration.
\newblock \emph{IEEE Transactions on Automation Science and Engineering}, 2025{\natexlab{d}}.
\newblock \doi{10.1109/TASE.2025.3560680}.

\bibitem[Zhou et~al.(2023)Zhou, Tian, Chu, Zhang, Zhang, Lu, Feng, Jie, Chiang, and Ma]{zhou2023fastpillars}
Zhou, S., Tian, Z., Chu, X., Zhang, X., Zhang, B., Lu, X., Feng, C., Jie, Z., Chiang, P.~Y., and Ma, L.
\newblock Fastpillars: A deployment-friendly pillar-based 3d detector.
\newblock \emph{arXiv preprint arXiv:2302.02367}, 2023.

\bibitem[Zhou et~al.(2024{\natexlab{a}})Zhou, Li, Zhang, Zhang, Bai, Sun, Zhao, Lu, and Chu]{zhou2024lidarptq}
Zhou, S., Li, L., Zhang, X., Zhang, B., Bai, S., Sun, M., Zhao, Z., Lu, X., and Chu, X.
\newblock Li{DAR}-{PTQ}: Post-training quantization for point cloud 3d object detection.
\newblock 2024{\natexlab{a}}.

\bibitem[Zhou et~al.(2024{\natexlab{b}})Zhou, Yuan, Yang, Zhao, Hu, Shi, Lu, and Wu]{zhou2024information}
Zhou, S., Yuan, Z., Yang, D., Zhao, Z., Hu, X., Shi, Y., Lu, X., and Wu, Q.
\newblock Information entropy guided height-aware histogram for quantization-friendly pillar feature encoder.
\newblock \emph{arXiv preprint arXiv:2405.18734}, 2024{\natexlab{b}}.

\bibitem[Zhou et~al.(2025{\natexlab{e}})Zhou, Nie, Zhao, Cao, and Lu]{lu2025focustrack}
Zhou, S., Nie, J., Zhao, Z., Cao, Y., and Lu, X.
\newblock Focustrack: One-stage focus-and-suppress framework for 3d point cloud object tracking.
\newblock In \emph{Proceedings of the 33rd ACM International Conference on Multimedia}, 2025{\natexlab{e}}.

\bibitem[Zhou et~al.(2025{\natexlab{f}})Zhou, Wang, Yuan, Shi, Shang, and Yang]{gsq}
Zhou, S., Wang, S., Yuan, Z., Shi, M., Shang, Y., and Yang, D.
\newblock {GSQ}-tuning: Group-shared exponents integer in fully quantized training for {LLM}s on-device fine-tuning.
\newblock In \emph{Findings of the Association for Computational Linguistics: ACL 2025}, pp.\  22971--22988, Vienna, Austria, July 2025{\natexlab{f}}. Association for Computational Linguistics.
\newblock ISBN 979-8-89176-256-5.
\newblock URL \url{https://aclanthology.org/2025.findings-acl.1178/}.

\bibitem[Zhou et~al.(2025{\natexlab{g}})Zhou, Yuan, Yang, Hu, Qian, and Zhao]{pillarhist}
Zhou, S., Yuan, Z., Yang, D., Hu, X., Qian, J., and Zhao, Z.
\newblock Pillarhist: A quantization-aware pillar feature encoder based on height-aware histogram.
\newblock In \emph{Proceedings of the Computer Vision and Pattern Recognition Conference}, pp.\  27336--27345, 2025{\natexlab{g}}.

\bibitem[Zuo et~al.(2025)Zuo, Li, Gong, and Tian]{11020157}
Zuo, H., Li, Z., Gong, J., and Tian, Z.
\newblock Intelligent road crack detection and analysis based on improved yolov8.
\newblock In \emph{2025 8th International Conference on Advanced Algorithms and Control Engineering (ICAACE)}, 2025.
\newblock \doi{10.1109/ICAACE65325.2025.11020157}.

\end{thebibliography}

\newpage
\appendix
\onecolumn


\end{document}